\documentclass[sigconf]{acmart}

\usepackage{soul}
\newcommand{\cal}[1]{\mathcal{#1}}
\usepackage{color}
\definecolor{Orange}{rgb}{0.9,0.5,0}
\definecolor{NavyBlue}{rgb}{0.1, 0.4, 0.8}
\definecolor{Magenta}{rgb}{0.8, 0.1, 0.6}
\definecolor{mypink2}{RGB}{219, 48, 122}

\usepackage{tabularx}
\usepackage{ragged2e}
\usepackage{multirow}
\newcommand{\specialcell}[2][l]{%
	\renewcommand{\arraystretch}{1.0} %
	\begin{tabular}[#1]{@{}l@{}}#2\end{tabular}}
\newcommand{\specialcellr}[2][r]{%
	\renewcommand{\arraystretch}{1.0} %
	\begin{tabular}[#1]{@{}r@{}}#2\end{tabular}}
\newcolumntype{M}{>{\RaggedRight\arraybackslash}X} 

\AtBeginDocument{%
	\providecommand\BibTeX{{%
			\normalfont B\kern-0.5em{\scshape i\kern-0.25em b}\kern-0.8em\TeX}}}

\copyrightyear{2020}
\acmYear{2020}
\setcopyright{acmlicensed}\acmConference[MM '20]{Proceedings of the 28th ACM International Conference on Multimedia}{October 12--16, 2020}{Seattle, WA, USA}
\acmBooktitle{Proceedings of the 28th ACM International Conference on Multimedia (MM '20), October 12--16, 2020, Seattle, WA, USA}
\acmPrice{15.00}
\acmDOI{10.1145/3394171.3413505}
\acmISBN{978-1-4503-7988-5/20/10}

\begin{document}
	\fancyhead{}
	
	\title{Describe What to Change: A Text-guided Unsupervised Image-to-Image Translation Approach}
	

	\author{Yahui Liu}
	\affiliation{%
		\institution{University of Trento, Italy}}
	\affiliation{\institution{FBK, Italy}}
	\email{yahui.liu@unitn.it}
	
	\author{Marco De Nadai}
	\affiliation{%
		\institution{FBK, Italy}}
	\email{denadai@fbk.eu}
	
	\author{Deng Cai}
	\affiliation{%
		\institution{The Chinese University of Hong Kong, China}}
	\email{thisisjcykcd@gmail.com}
	
	\author{Huayang Li}
	\affiliation{%
		\institution{Tencent AI Lab, China}}
	\email{alanili@tencent.com}
	
	\author{Xavier Alameda-Pineda}
	\affiliation{%
		\institution{Inria Grenoble Rh\^one-Alpes, France}}
	\email{xavier.alameda-pineda@inria.fr}
	
	\author{Nicu Sebe}
	\affiliation{%
		\institution{University of Trento, Italy}}
	\affiliation{\institution{Huawei Research, Ireland}}
	\email{niculae.sebe@unitn.it}
	
	\author{Bruno Lepri}
	\affiliation{\institution{FBK, Italy}}
	\email{lepri@fbk.eu}

	\renewcommand{\shortauthors}{Liu, et al.}
	\renewcommand{\shorttitle}{Describe What to Change}
	
	\begin{abstract}
		Manipulating visual attributes of images through human-written text is a very challenging task. On the one hand, models have to learn the manipulation without the ground truth of the desired output. 
		On the other hand, models have to deal with the inherent ambiguity of natural language. Previous research usually requires either the user to describe all the characteristics of the desired image or to use richly-annotated image captioning datasets. 
		In this work, we propose a novel unsupervised approach, based on image-to-image translation, that alters the attributes of a given image through a command-like sentence such as ``\textit{change the hair color to black}". Contrarily to state-of-the-art approaches, our model does not require a human-annotated dataset nor a textual description of all the attributes of the desired image, but only those that have to be modified.
		Our proposed model disentangles the image content from the visual attributes, and it learns to modify the latter using the textual description, before generating a new image from the content and the modified attribute representation.
		Because text might be inherently ambiguous (blond hair may refer to different shadows of blond, e.g. golden, icy, sandy), our method generates multiple stochastic versions of the same translation.
		Experiments show that the proposed model achieves promising performances on two large-scale public datasets: CelebA and CUB.  We believe our approach will pave the way to new avenues of research combining textual and speech commands with visual attributes. 
	\end{abstract}

	\vspace{-1em}

	
	\vspace{-1em}
	\vspace{-1em}
	
	
	\maketitle
	
	\vspace{-1em}
	\section{Introduction}
	Editing image attributes on portable devices is an uncomfortable operation for most people, but especially for impaired users. 
	On smartphones, changing the hair color on a picture requires to manually select the pixels to be altered and color them properly. 
	Inspired by the emergence of voice assistants, which 
	provide us a more convenient way to interact with machines through human language, we make pilot attempts to manipulate images through textual commands.
	Meeting this goal requires a deep integration of vision and language techniques, as previously achieved for other tasks, namely:
	image captioning~\cite{kaiser2017one, yao2017boosting}, text-to-image generation~\cite{li2019controllable,li2019manigan,nam2018text, Reed2016, zhang2017stackgan}, text-based video editing~\cite{fried2019text} and drawing-based image editing~\cite{jo2019sc, pathak2016context}. However, modifying parts of the image through natural language is still a challenging research problem.  
	
	\begin{figure*}[t]
		\vspace{-2mm}
		\includegraphics[width=\textwidth]{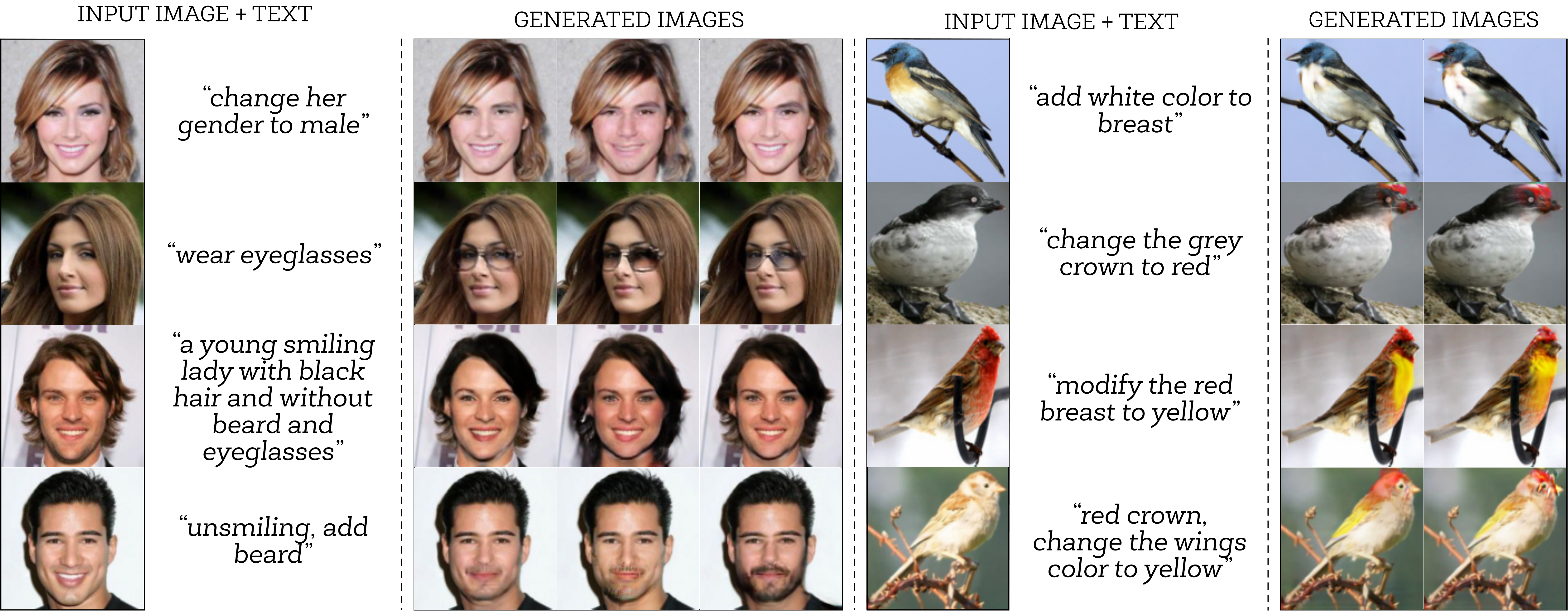}
		\vspace{-5mm}
		\captionof{figure}{Our model allows to manipulate visual attributes through human-written text. To deal with the inherent ambiguity of textual commands, our model generates multiple versions of the same translation being as such multi-modal. Here, we see some examples of generated images from the CelebA~\cite{liu2015deep} and CUB~\cite{wah2011caltech} datasets.
		}
		\label{fig:teaser}
	\end{figure*}
	
	Existing approaches, based on Generative Adversarial Networks (GANs)~\cite{arjovsky2017wasserstein, goodfellow2014generative}, require either to describe all the characteristics of the desired image~\cite{Reed2016, zou2019language} (e.g. ``\textit{young oval-faced lady with blond and black hair, smiling}"), or collecting a richly detailed dataset of supervised actions~\cite{el2018keep, el2019tell} or captions~\cite{nam2018text, Reed2016, reed2016learning, xu2018attngan, li2019manigan}.  
	For example, TAGAN~\cite{nam2018text} uses text input to change an image in order to have the attributes explained in the provided text description, but this method relies on detailed human-written captions. Similarly, El-Nouby \emph{et al.}~\cite{el2018keep, el2019tell} rely on annotated datasets to interact with the user and to generate semantically-consistent images from textual instructions.
	Two main issues have hindered the possibility to have an effective and unsupervised approach that follows human-written commands (e.g. ``\textit{make him smile}"). On the one hand, the literature on text-based image manipulation does not explicitly model image attributes, which would indicate if a specific part of the image is going to be changed or not. On the other hand, most GAN models assume a deterministic mapping, strongly limiting the capacity to handle the inherent ambiguity of human textual commands (e.g. ``\textit{blond hair}" might mean different shades of blond colors).

	In this paper, we propose \textit{Describe What to Change} (DWC), a novel method based on GANs to manipulate visual attributes from a human-written textual description of the required changes, thus named DWC-GAN. To explicitly model visual attributes and cope with the text ambiguity, our model translates images from one visual domain to other visual domains in a multi-modal (stochastic) fashion. Our contribution is three-fold:
	\begin{itemize}
		\item We propose the use of textual commands instead of image captions. 
		As shown in Table~\ref{tab:Comparisons_caption}, this leads to three advantages: (1) the commands can be more flexible than image captions and we can compose them for progressive manipulation; (2) the commands can be automatically generated, and (3) users are not required to know and mention all the modeled visual attributes in an input image, only the desired changes. DWC-GAN thus does not rely on human-annotated captioning text compared with state-of-the-art approaches~\cite{nam2018text, Reed2016, reed2016learning, li2019manigan}.
		\item Our model disentangles the content features and the attribute representation from the input image making the manipulation more controllable and robust. The attribute space follows a Gaussian Mixture Model (GMM), where the attributes of the source and the modified image follow the Gaussian distribution associated to their respective domains.
		\item To the best of our knowledge, we are the first to manipulate images through textual commands in a multi-modal fashion, allowing multiple versions of the same translation. 
	\end{itemize}
	
	\begin{table}[ht]
		\setlength{\tabcolsep}{5pt}
		\small
		\centering
		\begin{tabularx}{\columnwidth}{M M}
			\toprule
			\textbf{Input text} & \textbf{Features} \\
			\midrule
			\multirow{3}{\hsize}{\textbf{Captioning text:} \texttt{This woman is old and smiling, with black hair, eyeglasses and no beard.}} & - Enumerates all the attributes; \\
			& - No need to understand attributes from the input image; \\
			& - Often trained with positive and negative annotation for each translation pair~\cite{nam2018text,li2019manigan}. \\
			\midrule
			\multirow{3}{\hsize}{\textbf{Textual command:} \texttt{Change the hair to be black, increase her age.}} & - Focuses only on the differences; \\
			& - Requires the extraction of attributes from the input image; \\
			& - Trained without ground truth for each translation pair. \\
			\bottomrule
		\end{tabularx}
		\caption{Example of text describing a translation from an image of a young smiling woman with blond hair and eyeglasses to an older smiling woman with black hair and eyeglasses.
			Differently from captioning text, users are not required to know and mention all the modeled attributes. }
		\label{tab:Comparisons_caption}
		\vspace{-4mm}
	\end{table} 
	
	\section{Related Work}
	Our work is best placed in the literature of image-to-image translation and text-to-image generation. The former aims to transform an input image belonging to a visual domain (e.g. young people) to another domain (e.g. elderly people), while the latter tries to generate images from a textual description provided as input. 
	These two fields have witnessed many improvements in quality and realism, thanks to the advent of GANs~\cite{goodfellow2014generative, mirza2014conditional} and, in particular, conditional GANs (cGANs)~\cite{mirza2014conditional}, which are conditioned by extra information (e.g. the hair colour in face generation). GANs aim to synthesize images through a min-max game between a discriminator, trying to discriminate between real and fake data, and a generator, seeking to generate data that resemble the real ones. 
	
	\textbf{Image-to-image translation}: Conditional GANs were first employed for image-to-image translation in pix2pix~\cite{isola2017image} for learning a mapping from one domain to another by minimizing the $L_1$ loss between the generated and the target image. 
	However, pix2pix requires a large amount of paired data (e.g. colored-greyscale images), which is often unrealistic to collect. Thus, the interest had shifted to unsupervised models that learn to translate images with unpaired data, for which the image content might vary (e.g. daylight-night images can have different people and cars). These models require additional constraints to better differentiate the domain-dependent parts of the image from the domain-independent ones.
	For example, Liu \emph{et al.}~\cite{liu2015deep} assume that the two domains share a common latent space. CycleGAN~\cite{zhu2017unpaired} requires an image translated from a domain A to a domain B to be translated back to A, and applies a consistency loss, which was later employed by many other works~\cite{choi2018stargan, huang2018multimodal,  liu2019gesture, mo2018instanceaware}.
	MUNIT~\cite{huang2018multimodal} assumes that images share a domain-invariant content latent space and a style latent space, which is specific to a domain. This choice allows for generating multiple samples of the translation by drawing from the style distribution. Most of these approaches are, however, limited to one-to-one domain translation, which requires training a large number of models in the case of multiple domains.
	StarGAN~\cite{choi2018stargan} proposed a unified approach for multi-domain translation by employing a binary domain vector as input, which specifies the target domain, and a domain classification loss that helps the translation.
	Recently, GMM-UNIT~\cite{liu2020gmm} proposed a unified approach for multi-domain and multi-modal translation by modeling attributes through a Gaussian mixture, in which each Gaussian component represents a domain. In this paper, inspired by GMM-UNIT, we build a translation system where both the original and the manipulated attributes follow a GMM.
	
	\textbf{Text-to-image generation}: 
	cGANs can be conditioned with complex information, such as human-written descriptions, and then can generate accurate images, as shown by Reed \emph{et al.}~\cite{Reed2016, reed2016learning}. For example,  StackGAN~\cite{zhang2017stackgan} uses two stages of GANs in order to generate high-resolution images. The first stage generates a low-resolution image with the primitive shape and colour of the desired object, while the second one further refines the image and synthesize 256x256 images.
	The same authors improved the model with multiple generators through StackGAN++~\cite{zhang2018stackgan++}.
	Qiao \emph{et al.}~\cite{qiao2019mirrorgan} focused instead on semantic consistency, enforcing the generated images to have the same semantics with the input text description. In other words, the caption synthesized from the generated image should have the same meaning as the input text.
	Recently, Li \emph{et al.} proposed StoryGAN~\cite{li2019storygan}, which generates a series of images that are contextually coherent with previously generated images and with the sequence of text descriptions provided by the user. 
	
	These approaches, however, require to accurately describe the picture to be generated, without allowing to start from an existing image and modify it through text.
	
	\textbf{Image manipulation}: Modifying images through some user-defined conditions is a challenging task. Most of the previous approaches rely on conditional inpaiting~\cite{bau2019semantic, jo2019sc, zheng2019pluralistic}, in which the network fills a user-selected area with pixel values coherent with the user preferences and the context. 
	However, image editing does not necessarily require to select the exact pixels that have to be changed.
	Wang \emph{et al.}~\cite{wang2018learning} learned to change global image characteristics such as brightness and white balance from textual information.
	Zou \emph{et al.}~\cite{zou2019language} proposed a network that colourizes sketches following the instructions provided by the input text specifications. 
	El-Nouby \emph{et al.}~\cite{el2018keep, el2019tell} introduced a network that generates images from a dialogue with the user. A first image is generated from text, then objects are added according to the user preference provided as free text.
	Chen \emph{et al.}~\cite{chen2018language} learned with a recurrent model and an attention module to colourize greyscale images from a description of the colours or the scene (e.g. \textit{``the afternoon light flooded the little room from the window"}). 
	Recently, Cheng \emph{et al.}~\cite{cheng2018sequential} proposed a network, tailored for dialogue systems, that gets as inputs an image and a series of textual descriptions that are apt to modify the image. The network is encouraged to synthesize images that are similar to the input and the previously generated images. However, examples show that the content of the image might vary substantially.
	Nam \emph{et al.}~\cite{nam2018text} and Li \emph{et al.}~\cite{li2019manigan} introduced instead TAGAN and ManiGAN that modify fine-grained attributes such as the colours of birds' wings through text. The key idea of their methods is reconstructing the images with positive and negative sentences, where the positive/negative refers to captioning text matching/mismatching the corresponding image. 
	The recurrent network is trained to learn the words that refer to the attributes in the images and it allows to change words to manipulate multiple attributes at once.
	
	Current image manipulation methods based on text often rely on human-annotated captioning text~\cite{nam2018text, Reed2016, reed2016learning, xu2018attngan, li2019manigan}, which describes the scene in the image. Moreover, they do not explicitly model the concept of attributes (i.e. domains), sometimes failing at balancing the trade-off between keeping the old content and changing it~\cite{nam2018text}, which is a well-known issue in image-to-image translation. Last but not least, existing solutions are deterministic, thus limiting the diversity of the translated outputs, even if the provided text might have different meanings. For example, ``\textit{wear a beard}" might mean to have a goatee, short or long beard.
	
	Therefore, we build on state-of-the-art approaches for image-to-image translation and on the image manipulation literature and propose to explicitly model the visual attributes of an image and learn how to translate them through automatically generated commands. We are, up to the best of our knowledge, the first proposing a multi-domain and multi-modal method that translates images from one domain to another one by means of the user descriptions of the attributes to be changed. 
	
	\begin{figure*}[!ht]
		\centering
		\includegraphics[width=0.8\textwidth]{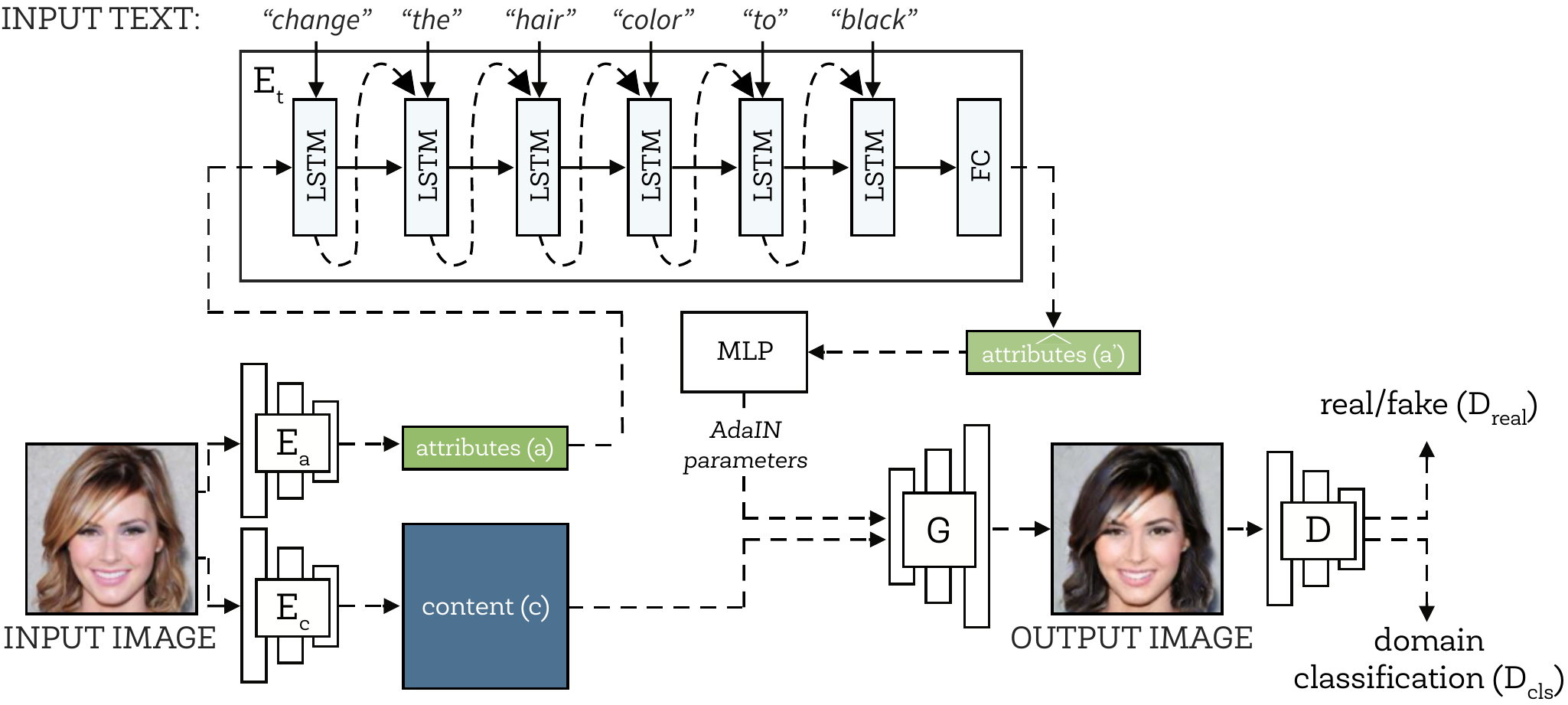}
		\vspace{-4mm}
		\caption{Architecture of our model. First, we disentangle the attributes and the content of the input image. Then, we modify the visual attributes of the original image using a text encoder. The generator uses a MLP to produce a set of AdaIN~\cite{huang2017arbitrary} parameters from the attribute representation. The content features is then processed by the parameterized generator to predict a realistic image with required attributes. Finally, the discriminator classifies whether the generated image is real or fake.}
		\label{fig:framework}
		\vspace{-4mm}
	\end{figure*}
	
	\section{Methodology}
	
	Our model is designed to modify the attributes of an image $\mathbf{x}$ through a human-written text $\mathbf{t}$, which describes the changes to be made in the image. In other words, we want to generate a new image $\hat{\mathbf{x}}$ that has the visual content of $\mathbf{x}$ and all its attributes but those attributes that should be altered accordingly to the text $\mathbf{t}$.
	Different from previous works, $\mathbf{t}$ is not a description of all the attributes of the generated image, but it is a direct command describing the modifications to be made (e.g. \textit{change the hair color to black}). 
	Thus, the proposed method should be able to model the attributes of the original image, the modifications of (some of) the attributes described in the text, and the attributes of the generated image. Moreover, it has to deal with the ambiguity of input text, which does not describe all the attributes of the target image, nor it is direct and clear as one-hot vector inputs. 
	
	The architecture, as shown in Figure~\ref{fig:framework}, processes the input as follows. Given $N$ attributes of a set of images, we first disentangle the visual content $\mathbf{c}\in\pmb{\mathcal{C}}\subset\mathbb{R}^C$ and the attributes $\mathbf{a}\in\pmb{\mathcal{A}}\subset\mathbb{R}^N$ of the input image $\mathbf{x}$ through the encoders $E_c$ and $E_a$, respectively. Then, $E_t$, which is a Recurrent Neural Network (RNN) module, uses the input sentence $\mathbf{t}$ and the extracted attributes $\mathbf{a}$ to infer the desired (target) attributes $\mathbf{a}'=E_t(\mathbf{t},\mathbf{a})$. The inferred target attributes are then used along with the extracted content $\mathbf{c}$, by the generator $G$, to output the image $G(\mathbf{c},\mathbf{a}')$. Finally, the discriminator $D$ discerns between ``real'' or ``fake'' images ($D_{real}$) and recognizes the domain of the generated image ($D_{cls}$). Here, a domain is a combination of  visual attributes.  The model (content and attribute extractors, generator and discriminators) is learned in an end-to-end manner.
	
	\vspace{-1em}
	\subsection{Assumptions}
	\label{subsec:assumptions}
	
	The basic assumption of DWC-GAN is that each image can be decomposed in a domain-invariant content space and a domain-specific attribute space~\cite{gonzalez2018image, huang2018multimodal,liu2020gmm}. We let the network model the high-dimension content features, while we represent the attributes of the image through Gaussian components in a mixture. This representation allows to model the combinations of attributes in a continuous space. We can then exploit this space to work with attribute combinations that have never (or little) been observed in the data. Formally and similarly to GMM-UNIT~\cite{liu2020gmm}, we model the attributes with a $K$-component $d$-dimensional GMM: $p(\mathbf{a}) = \sum_{k=1}^{K}\phi_k{\cal N}(\mathbf{a}; \pmb{\mu}^k, \pmb{\Sigma}^k)$
	where $\phi_k$, $\pmb{\mu}^k$ and $\pmb{\Sigma}^k$ denote respectively the weight, mean vector and covariance matrix of the $k$-th GMM component ($\phi_k\geq 0$, $\sum_{k=1}^K \phi_k = 1$, $\pmb{\mu}^k\in\mathbb{R}^d$ and $\pmb{\Sigma}^k\in\mathbb{R}^{d\times d}$ is symmetric and positive definite). 
	Therefore, we model the attributes in a domain through a corresponding GMM component. Thus, for an image $\mathbf{x}$ from domain $\pmb{\mathcal{X}}^k$ (i.e. $\mathbf{x}\sim p_{\pmb{\mathcal{X}}^k}$), its latent attribute $\mathbf{a}$ is assumed to follow the $k$-th Gaussian component: $\mathbf{a}^k\sim{\cal N}(\pmb{\mu}^k, \pmb{\Sigma}^k)$.
	We aim to manipulate a source image $\mathbf{x}$ from $\pmb{\mathcal{X}}^{k}$ into domain $\pmb{\mathcal{X}}^{\ell}$ through a textual command $\mathbf{t}^{\ell}\sim p_{\pmb{\mathcal{T}}^{\ell}}$, where $k,\ell \in \{1,\dots,K\}$. 
	
	\subsection{Multi-modal image generation}
	In order to have a multi-modal (non-deterministic) translation, we enforce both the sampled and extracted attributes to follow the image domain distribution, that is $\forall \ell, k$:
	\begin{equation*}
	\begin{aligned}
	& G(E_c(\mathbf{x}),\mathbf{a})\sim p_{\pmb{\mathcal{X}}^k}, \quad \forall\; \mathbf{a}\sim {\cal N}(\pmb{\mu}^k, \pmb{\Sigma}^k), \mathbf{x}\sim p_{\pmb{\mathcal{X}}^{\ell}}.\\
	\end{aligned}
	\end{equation*}
	We ensure these two properties with the following losses.
	
	\noindent \textbf{Reconstruction losses.} These losses, introduced at first in~\cite{huang2018multimodal, zhu2017unpaired}, force the output to be consistent with the original content and attributes. Specifically, the \emph{self-reconstruction loss} ensures that the original image is recovered if its attribute and content is used in the generator. If $\mathbf{c}_{\mathbf{x}} = E_c(\mathbf{x})$ and $\mathbf{a}_{\mathbf{x}} = E_a(\mathbf{x})$, then for all $k$:
	\begin{equation*}
	\mathcal{L}_{rec\_s} =  
	\mathbb{E}_{ \mathbf{x} \sim p_{\pmb{\mathcal{X}}^k}} \big[\|G(\mathbf{c}_{\mathbf{x}}, \mathbf{a}_{\mathbf{x}}) - \mathbf{x}\|_1 \big] .
	\end{equation*}
	The \emph{content and attribute reconstruction losses}~\cite{huang2018multimodal} are used to constrain and learn the content and attribute extractors: 
	\begin{equation*}
	\mathcal{L}_{rec\_c} = 
	\mathbb{E}_{\mathbf{x} \sim p_{\pmb{\mathcal{X}}^k}, \mathbf{a} \sim {\cal N}(\pmb{\mu}^{\ell}, \pmb{\Sigma}^{\ell})} \big[\| E_c(G(\mathbf{c}_{\mathbf{x}}, \mathbf{a})) - \mathbf{c}_{\mathbf{x}}\|_1 \big],
	\end{equation*}
	\begin{equation*}
	\mathcal{L}_{rec\_a} = 
	\mathbb{E}_{\mathbf{x} \sim p_{\pmb{\mathcal{X}}^k}} \big[\| E_a(G(\mathbf{c}_{\mathbf{x}}, \mathbf{a}_\mathbf{x})) - \mathbf{a}_\mathbf{x}\|_1 \big].
	\end{equation*}
	The \emph{cycle reconstruction} loss~\cite{zhu2017unpaired} enforces consistency when translating an image into a new domain and then back to the original one:
	\begin{equation*}
	\begin{aligned}
	\mathcal{L}_{cyc} = & 
	\mathbb{E}_{\mathbf{x} \sim p_{\pmb{\mathcal{X}}^k},\mathbf{a} \sim {\cal N}(\pmb{\mu}^{\ell},\pmb{\Sigma}^{\ell})}
	\big[\|G(E_c(G(\mathbf{c}_{\mathbf{x}}, \mathbf{a})), \mathbf{a}_{\mathbf{x}}) - \mathbf{x}\|_1\big],
	\end{aligned}
	\end{equation*}
	where the use of the $\mathcal{L}_1$ loss inside the expectation is motivated by previous works~\cite{isola2017image} showing that $\mathcal{L}_1$ produces sharper results than $\mathcal{L}_2$. To encourage the generator to produce diverse images, we explicitly regularize $G$ with the \textit{diversity sensitive loss}~\cite{mao2019mode,choi2019stargan}:
	\begin{equation*}
	\mathcal{L}_{ds} =  \mathbb{E}_{\mathbf{x} \sim p_{\pmb{\mathcal{X}}^k}, \mathbf{a}_1, \mathbf{a}_2 \sim {\cal N}(\pmb{\mu}^{\ell}, \pmb{\Sigma}^{\ell})} \big[\| G(\mathbf{c}_{\mathbf{x}}, \mathbf{a}_1) - G(\mathbf{c}_{\mathbf{x}}, \mathbf{a}_2)\|_1 \big]
	\end{equation*}
	
	\noindent \textbf{Domain losses.} Similarly to StarGAN~\cite{choi2018stargan}, for any given input image $\mathbf{x}$, we would like the method to classify it as its original domain, and to be able to generate an image in any domain from its content. Therefore, we need two different losses, one directly applied to the original images, and a second one applied to the generated images:
	\begin{equation*}
	\mathcal{L}_{cls}^D = \mathbb{E}_{\mathbf{x}\sim p_{\pmb{\mathcal{X}}^k}, d_{\pmb{\mathcal{X}}^k}}[-\log D_{cls}(d_{\pmb{\mathcal{X}}^k}|\mathbf{x})] \quad \text{and}
	\end{equation*}
	\begin{equation*}
	\begin{aligned}
	\mathcal{L}_{cls}^G = & 
	\mathbb{E}_{\mathbf{x} \sim p_{\pmb{\mathcal{X}}^k}, d_{\pmb{\mathcal{X}}^{\ell}}, \mathbf{a}\sim {\cal N}(\pmb{\mu}^{\ell},\pmb{\Sigma}^{\ell})}[ -\log D_{cls}(d_{\pmb{\mathcal{X}}^{\ell}}|G(\mathbf{c}_{\mathbf{x}}, \mathbf{a}))],
	\end{aligned}
	\end{equation*}
	where $d_{\pmb{\mathcal{X}}^\ell}$ is the label of $\ell$-th domain. Importantly, the discriminator $D$ is trained using the first loss, while the generator $G$ is trained using the second loss.
	
	\noindent \textbf{Adversarial losses.} These terms enforce the generated images to be indistinguishable from the real images by following the formulation of LSGAN~\cite{mao2017least}:
	\begin{equation*}
	\begin{aligned}
	\mathcal{L}_{GAN}^D =  
	& \mathbb{E}_{\mathbf{x}\sim p_{\pmb{\mathcal{X}}^k}}[ D_{\text{real}}(\mathbf{x})^2]  +  \\ & \mathbb{E}_{\mathbf{x} \sim p_{\pmb{\mathcal{X}}^k}, \mathbf{a}\sim {\cal N}(\pmb{\mu}^{\ell},\pmb{\Sigma}^{\ell})}[(1-D_{\text{real}}(G(\mathbf{c}_{\mathbf{x}}, \mathbf{a})))^2]
	\end{aligned}
	\end{equation*}
	\begin{equation*}
	\mathcal{L}_{GAN}^G  = 
	\mathbb{E}_{\mathbf{x} \sim p_{\pmb{\mathcal{X}}^k}, \mathbf{a} \sim {\cal N}(\pmb{\mu}^{\ell},\pmb{\Sigma}^{\ell})}[D_{\text{real}}(G(\mathbf{c}_{\mathbf{x}}, \mathbf{a}))^2]
	\end{equation*}
	
	\subsection{Attribute manipulation losses}
	As mentioned above, the underlying statistical assumption for the attributes is to follow a GMM -- one component per domain. Very importantly, both the attribute representations extracted from image by $E_a$
	and the attribute representations obtained from $E_t$ should follow the assumption and correspond to the correct component.
	We do that by imposing that the Kullback–Leibler (KL) divergence  of the extracted and manipulated attributes correspond to the Gaussian component of the original and targetted domain respectively. Recalling that $\mathbf{a}_{\mathbf{x}} = E_a(\mathbf{x})$, we write:
	\begin{equation*}
	\begin{aligned}
	\mathcal{L}_{KL} = \mathbb{E}_{\mathbf{x}\sim p_{\pmb{\mathcal{X}}^k},  \mathbf{t} \sim p_{\pmb{\mathcal{T}}^{\ell}}}  \big[ &   \mathcal{D}_{KL}(\mathbf{a}_{\mathbf{x}}\|\mathcal{N}(\pmb{\mu}^k, \pmb{\Sigma}^k)) + \\ & \mathcal{D}_{KL}(E_t(\mathbf{t},\mathbf{a}_{\mathbf{x}})\|\mathcal{N}(\pmb{\mu}^{\ell}, \pmb{\Sigma}^{\ell}))\big] 
	\label{eq:kl}
	\end{aligned}
	\end{equation*}
	where $\mathcal{D}_{KL}(p\|q) = -\int p(t)\log\frac{p(t)}{q(t)}dt$ is the KL.
	
	Intuitively, the second KL divergence enforces the images generated from the manipulated attributes to follow the distribution of the target domain:
	\begin{equation*}
	\begin{aligned}
	& G(\mathbf{c}_\mathbf{x},E_t(\mathbf{t}, \mathbf{a}_\mathbf{x}))\sim p_{\pmb{\mathcal{X}}^{\ell}}, \quad \forall \;\mathbf{x}\sim p_{\pmb{\mathcal{X}}^k}, \mathbf{t} \sim p_{\pmb{\mathcal{T}}^{\ell}}
	\end{aligned}
	\end{equation*}
	
	Finally, the full objective function of our network is:
	\begin{equation*}
	\mathcal{L}_D = \mathcal{L}_{GAN}^D + \lambda_{cls}\mathcal{L}_{cls}^D
	\end{equation*}
	\begin{equation*}
	\begin{aligned}
	\mathcal{L}_G = & \mathcal{L}_{{GAN}}^G + \lambda_{rec\_s} \mathcal{L}_{rec\_s} + \mathcal{L}_{rec\_c} + \mathcal{L}_{rec\_a} +  \\ & \lambda_{cyc}\mathcal{L}_{cyc} - \mathcal{L}_{ds} +  \lambda_{KL}\mathcal{L}_{KL} + \mathcal{L}_{cls}^G \\
	\end{aligned}
	\end{equation*}
	where $\{ \lambda_{rec\_s}, \lambda_{cyc}, \lambda_{KL}\}$ are hyper-parameters of weights for corresponding loss terms. The value of most of these parameters come from the literature. We refer to Supplementary Material for the details. Note that these losses are required to constrain down the difficult problem of image to image translation, disentangle content and style, and achieve stochastic results by sampling from the GMM distribution of attributes. Without these losses, the problem would be much less constrained and difficult.
	
	\subsection{Domain Sampling}
	We propose a strategy to obtain diverse manipulated results during testing by following the assumption introduced in Section~\ref{subsec:assumptions}. After extracting the target attributes $\mathbf{a}'$ from the original image $\mathbf{x}$ and input text $\mathbf{t}$, we assign it to a closest component $(\pmb{\mu}^*, \pmb{\Sigma}^*)$ of the GMM by following: $k^* = \mathop{\arg\max}_{k\in\{1,\ldots,K\}} \{\Phi (\mathbf{a}'; \pmb{\mu}^k, \pmb{\Sigma}^k)\} $, 
	where $\Phi$ is the Gaussian probability density function. Then, we can randomly sample several $\mathbf{a}^*$ from the component $\mathcal{N}(\pmb{\mu}^{k^*}, \pmb{\Sigma}^{k^*})$. With such different sampled $\mathbf{a}^*$'s and the trained non-linear generator $G$, our model achieves multi-modality in any domain, which is different from state-of-the-art approaches~\cite{nam2018text, zhang2018stackgan++} that use random noise sampled from a standard normal distribution $\mathcal{N}(0, 1)$ to generate multiple diverse results independently of the domain. 
	
	Sampling all the target attributes might hurt translation performance. Thus, we propose an additional constraint that the ideal attribute representation is a mixture of domain sampling $\mathbf{a}^*$ and extracted representation $E_a(\mathbf{\mathbf{x}})$, where $\mathbf{a}^*$ copies the attribute representation from $E_a(\mathbf{\mathbf{x}})$ when a target attribute is the same as the attribute in the input image.
	
	\subsection{Unsupervised Attention} 
	Attention mechanisms have proven successful in preserving the visual information that should not be modified in an image-to-image translation~\cite{mejjati2018unsupervised,pumarola2018ganimation,liu2020gmm,liu2019gesture}. For example, GMM-UNIT~\cite{liu2020gmm} claims that models are inclined to manipulate the intensity and details of pixels that are not related to the desired attribute transformation. Therefore, we follow the state-of-the-art approaches and add a convolutional layer followed by a sigmoid layer at the end of $G$ to learn a single channel attention mask $\mathbf{M}$ in an unsupervised manner. Then, the final
	prediction $\tilde{\mathbf{x}}$ is obtained from a convex combination of the input and the initial translation: 
	$\tilde{\mathbf{x}}=\hat{\mathbf{x}} \cdot \mathbf{M} + \mathbf{x}\cdot (1-\mathbf{M})$.
	
	\section{Experimental Setup}
	\subsection{Datasets}
	
	\noindent\textbf{CelebA.} 
	The CelebFaces Attributes (CelebA) dataset~\cite{liu2015deep} contains 202,599 face images of celebrities where each face is annotated with 40 binary attributes. 
	This dataset is composed of some attributes that are mutually exclusive (e.g. either male or female) and those that are mutually inclusive (e.g. people could have both blond and black hair). 
	To be consistent with previous papers~\cite{choi2018stargan, liu2020gmm}, we select a subset of attributes, namely \textit{black hair}, \textit{blond hair}, \textit{brown hair}, \textit{male/female}, \textit{young/old}, \textit{smile}, \textit{eyeglasses}, \textit{beard}. 
	As we model these attributes as different GMM components, our model allows the generation of images with up to $2^8$ different combinations of the attributes. Note that in the dataset only 100 combinations exist. As preprocessing, we center crop the initial 178$\times$218 images to 178$\times$178 and resize them to 128$\times$128. We randomly select 2,000 images for testing and use all remaining images for training.

	\noindent\textbf{CUB Birds.} The CUB dataset~\cite{wah2011caltech} contains 11,788 images of 200 bird species, and provides hundreds of annotated attributes for each bird (e.g.\ color, shape).
	We select some color attributes \{\textit{grey}, \textit{brown}, \textit{red}, \textit{yellow}, \textit{white}, \textit{buff}\} of three body parts \{\textit{crown}, \textit{wing}, \textit{breast}\} of the birds, yielding a total of $K=18$ attributes used to represent around one thousand domains appearing in the dataset.
	Differently from the CelebA dataset, the selected colors could appear in all selected body parts. Hence, the colors and body parts follow a hierarchical structure, requiring the model to modify the color of a specific body part. 
	We crop all images by using the annotated bounding boxes of the birds and resize images to 128$\times$128.
	
	Additional details can be found in the Supplementary Material.
	
	\subsection{Automatic Text Description}
	\label{sec:auto-text}
	The text encoder $E_t$ has to be trained with a dataset describing the changes to be made in the visual attributes of the input image. Therefore, we here propose a protocol to generate the corpus automatically for the datasets provided multiple annotated attributes or labels. Given an image $\mathbf{x}$, we can collect a set of attributes $\{p_1, \dots, p_N\}$ and obtain the corresponding real-valued vector $\mathbf{p}\in \{0,1\}^N$ for generating the description text. 
	We denote $\mathbf{p}'$ as the randomly assigned real-valued vector of a target image. After observing the human-used commands of manipulating images, we propose an automatic description text generation method through three different strategies:
	
	\noindent\textbf{Step-by-step actions.} We can compare the differences between $\mathbf{p}$ and $\mathbf{p}'$ element-by-element. For example, if the hair colors in the source image and target image are ``\textit{brown}" and ``\textit{black}", respectively. We can describe the action like ``\textit{change the brown hair to \ul{black}}", ``\textit{change the hair color to \ul{black}"}, ``\textit{\ul{black hair}}". Similarly, we can describe all actions that change the image one by one. For the same attributes in $\mathbf{p}$ and $\mathbf{p}'$, we can provide empty description or the no-change commands (e.g. ``\textit{do not change the hair color}", ``\textit{keep his hair color unchanged}"). Then, the sequence of step-by-step actions are disrupted randomly. 
	A full example for this strategy is ``\textit{make the face \ul{older}, translate the face to be \ul{smiling}, \ul{remove the eyeglasses}, do nothing on the \ul{gender}, change the hair color to \ul{black}, \ul{wear a beard}}".
	
	\noindent\textbf{Overall description.} In practice, it's feasible to describe the attributes in the target images to manipulate images, which is similar to the text-to-image task~\cite{nam2018text, Reed2016,reed2016learning,li2019manigan}. Here, we completely describe the appeared attributes in $\mathbf{p}'$. For example, ``\textit{a \ul{smiling} \ul{young} \ul{man} with \ul{black hair}, \ul{wearing a beard} and \ul{without eyeglasses}"}. In addition, the untouched attributes are not regarded as existing attributes in the target image by default.  
	
	\noindent\textbf{Mixed description.} To increase the generalization, we mix the two previous strategies by randomly mentioning some attributes in $\mathbf{p}$ and specifying the exact change of attributes in $\mathbf{p}'$
	For example, \textit{translate the young miss to be an \ul{old} \ul{smiling} \ul{man} with \ul{black hair} and wearing \ul{beards} and \ul{eyeglasses}."}. In this case, the attributes of both the input image and input text are mentioned, which is more complex for the language model.
	
	During training, these three strategies are randomly selected for each input image, and are used for both the CelebA and the CUB Birds datasets. More details are provided in the supplementary material.
	This dataset is released along with the paper to allow future work and better comparisons with our results.
	
	\subsection{Metrics}
	We quantitatively evaluate our model through the image quality, diversity and the accuracy of generated images. 
	We evaluate the image quality through the Fr\'echet Inception Distance (FID)~\cite{NIPS2017_7240} and the Inception Score (IS)~\cite{salimans2016improved}, the diversity is measured by the Learned Perceptual Image Patch Similarity (LPIPS)~\cite{zhang2018unreasonable}, the mean accuracy (Mean Acc) of multi-label classification, while we also measure the accuracy through a user study. 
	
	\noindent\textbf{FID and IS.} We randomly select 10000 source images and 10000 target images. The models transfer all source images with the same attributes of target images through textual commands, as shown in Section~\ref{sec:auto-text}. Hence, the FID is estimated using 10000 generated images vs the selected target 10000 images. The IS is evaluated using Inception-v3~\cite{szegedy2016rethinking} and 10000 generated images for FID.
	
	\noindent\textbf{LPIPS.} The LPIPS distance is defined as the $L_2$ distance between the features extracted by a deep learning model of two images. This distance has been demonstrated to match well the human perceptual similarity~\cite{zhang2018unreasonable}. Thus, following~\cite{ huang2018multimodal, zhu2017toward}, we randomly select 100 input images and translate them to different domains. For each domain translation, we generate 10 images for each input image and evaluate the average LPIPS distance between the 10 generated images. Finally, we get the average of all distances. Higher LPIPS distance indicates better diversity among the generated images. We show in the results both the mean and standard deviation.
	
	
	\noindent \textbf{Accuracy and Realism.} Through these two metrics we evaluate how humans perceive generated images. \textit{Accuracy} measures whether the attributes of the generated images are coherent with the target attributes (0: incorrect; 1: correct). \textit{Realism} quantifies how realistic a generated image is perceived (0: bad, unrecognizable; 1: neutral, recognizable but with obvious defects; 2: good, recognizable but with slight defects; 3: perfect, recognizable and without any defect). We test Accuracy and Realism by randomly sampling 50 images with the same translation condition and collect more than 20 surveys from different people with various background.
	
	\subsection{Baseline Models}
	
	\noindent\textbf{StackGAN++}~\cite{zhang2018stackgan++} is a text-to-image model that takes text as input and encodes the text into embedding for the decoding network. Here, we train/test this model by using the embedding of the manipulation text as input.
	
	\noindent\textbf{TAGAN}~\cite{nam2018text} 
	and \textbf{ManiGAN}~\cite{li2019manigan} 
	are two image-conditioned text-to-image models that take the original image and detailed description text of the target image as input. We follow the original method and train/test the model on our automatic description text.
	
	\noindent\textbf{StarGAN}~\cite{choi2018stargan} is a unified multi-domain translation model that takes the original image and target attribute vector as input. Here, we add the same RNN module (i.e. the text understanding module) used in our framework to StarGAN to extract attributes from the input text. After that, we feed the embedding of the input text and original image as input. We call this modified version as StarGAN*. 
	
	For all the baseline models, we use the same pre-trained word embeddings in 300 dimensions for the input description text. The embeddings are trained by the skip-gram model described in~\cite{bojanowski2017enriching} on Wikipedia using fastText\footnote[1]{\url{https://github.com/facebookresearch/fastText}}. 
	
	\begin{table*}[ht]
		\setlength{\tabcolsep}{5pt}
		\small
		\centering
		\begin{tabular}{@{}lrrr rr rr@{}}
			\toprule
			\textbf{Method} & {\bf IS}$\uparrow$ & {\bf FID}$\downarrow$ & {\bf LPIPS}$\uparrow$ & {\bf Accuracy}$\uparrow$ & {\bf Realism}$\uparrow$ & \textbf{Params~$\downarrow$} & \textbf{Training Speed (s/iter)}~$\downarrow$ \\
			\midrule
			StackGAN++~\cite{zhang2018stackgan++} & 1.444 & 285.48 & \textbf{.292}$\pm$.053 &  .000 & .00 & 50.30M & \textbf{.082}$\pm$.003\\
			TAGAN~\cite{nam2018text}  & 1.178 & 421.84 & .024$\pm$.012 &  .000 & .00 & 44.20M & .804$\pm$.001\\
			StarGAN*~\cite{choi2018stargan} & 2.541 & 50.66 & .000$\pm$.000 & .256 & 1.17 & 54.10M & .130$\pm$.001\\
			DWC-GAN (proposed) & \textbf{3.069} & \textbf{32.14} & .152$\pm$.003 & {\bf.885} & {\bf2.25} & \textbf{32.75}M & .172$\pm$.010 \\
			\bottomrule
		\end{tabular}
		\caption{Quantitative evaluation on the CelebA dataset. There is no captioning text on CelebA.}
		\label{tab:QuantitativeResults_ce}
		\vspace{-4mm}
	\end{table*} 
	
	\begin{figure}[!ht]
		\centering
		\includegraphics[width=\linewidth]{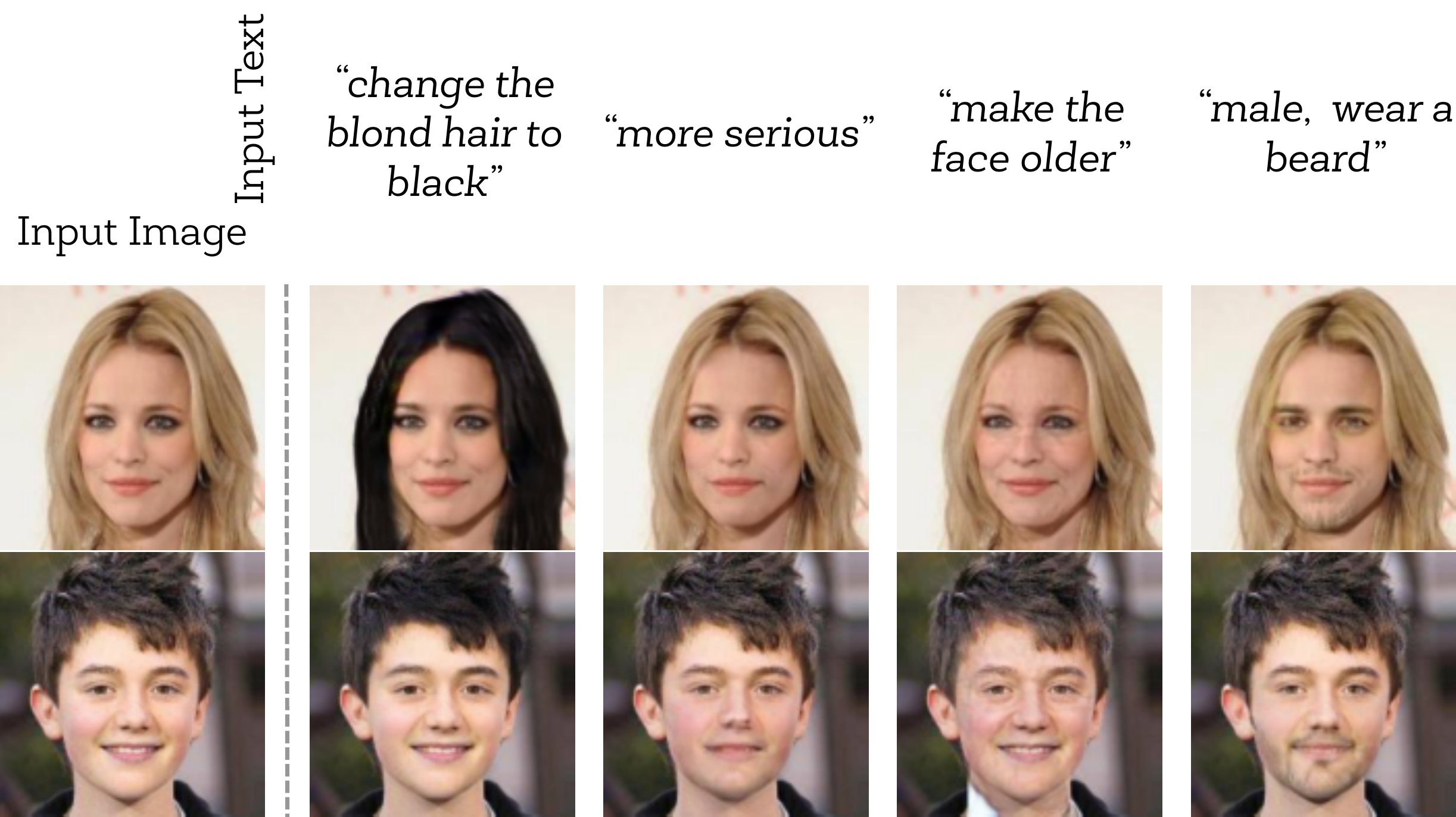}
		\vspace{-5mm}
		\caption{Qualitative evaluation for different textual input on CelebA dataset. Our model generates high-quality images that are consistent with the textual commands. 
		}
		\label{fig:visual_celeba}
		\vspace{-2mm}
	\end{figure}
	
	\begin{figure}[!ht]
		\centering
		\includegraphics[width=\linewidth]{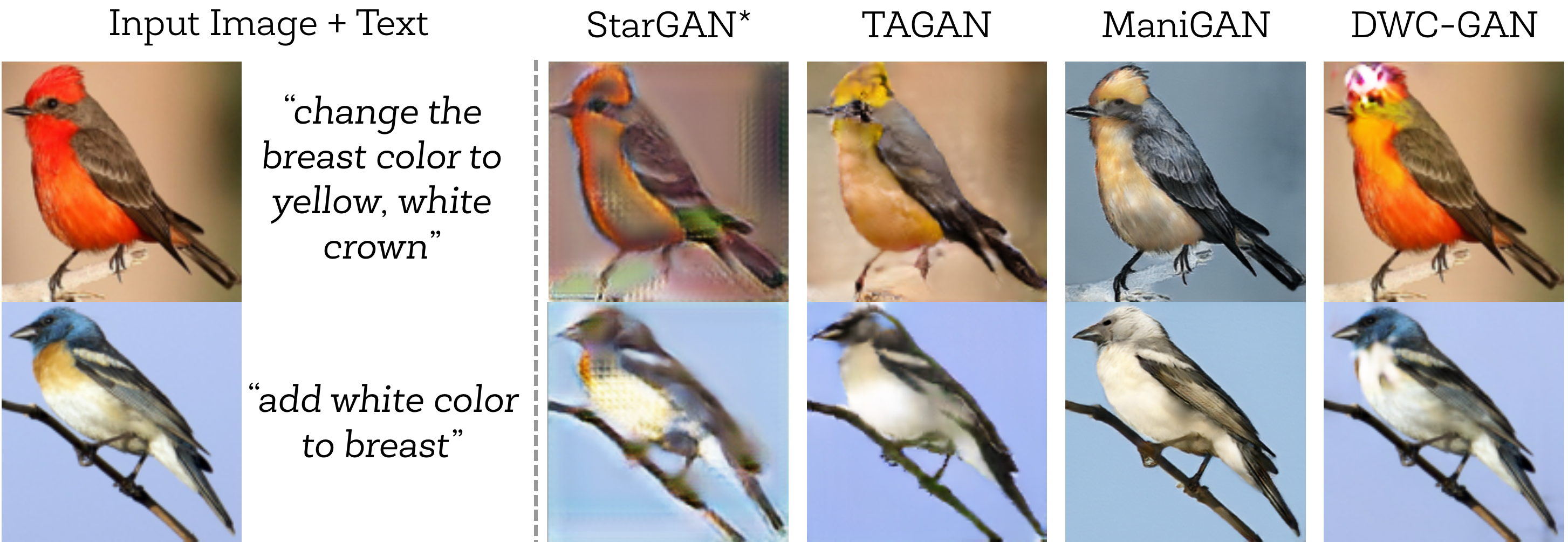}
		\vspace{-5mm}
		\caption{Qualitative comparisons for different textual input on CUB datasets. For reference, we show also the results of StarGAN*~\cite{choi2018stargan}, TAGAN~\cite{nam2018text} and ManiGAN~\cite{li2019manigan}. 
		}
		\label{fig:visual_cub200}
		\vspace{-2mm}
	\end{figure}
	
	\section{Results}
	We begin by quantitatively comparing our model with state of the art. Table \ref{tab:QuantitativeResults_ce} shows that our model generates better quality images than all competing methods, using around half the parameters and achieving comparable training speed. 
	Specifically, DWC-GAN outperforms our most similar competitor, TAGAN, attaining higher IS (3.069 vs 1.178) and lower FID (32.14 vs 421.84) with less required parameters (32.75M vs 44.20M). 
	Surprisingly, the diversity of generated images is lower than StackGAN++. However, the qualitative results show that StackGAN++ generates diverse but very noisy images, in which the attributes are not easily understandable (see the Supplementary Material for details).
	Contrarily, Figure~\ref{fig:visual_celeba} shows that DWC-GAN generates high-quality images that are also consistent with the manipulation described by the text. 
	Our approach outperforms state of the art in all the metrics in the CUB dataset as well, as shown in both Figure~\ref{fig:visual_celeba} and Table~\ref{tab:QuantitativeResults_cub}. Additional results can be seen in Figure~\ref{fig:teaser} and in the Supplementary Material.
	
	Differently from state of the art, DWC-GAN explicitly models the attributes of images in a smooth and continuous space through a GMM. This allows to translate images in multiple domains, but also to have multi-modal results. As Figure~\ref{fig:teaser} shows, our model generates different realizations of the same manipulation. For example, in \textit{``add beards"} the model synthesizes various versions of the subjects' beards, as the command is open to more than one interpretation. State-of-the-art models such as TAGAN and StarGAN* do instead generate results with no, or low, diversity (see the Supplementary Material for additional details).
	
	Our model uses the source attributes and the human-written sentence to generate the desired attributes. The results of StackGAN++ show that it is not feasible to only encode textual commands to generate meaningful target images, while StarGAN* shows that taking into account only the text attributes as an additional condition is not enough. In particular, StarGAN* exhibits a visible mode collapse in this setting. These results show that applying and modifying the existing baselines is not feasible to solve our task.
	
	We trained our model with automatically generated text, thus without relying on human-annotated datasets as state of the art does~\cite{nam2018text, Reed2016, reed2016learning}. We observe that while TAGAN does not perform well with our corpus, we have consistent results with different strategies of automatic text generation, and even with a text strategy that is similar to their dataset (the Overall strategy). 
	This result proves that annotating the dataset with automatically generated textual commands is a feasible strategy to train a model that manipulates images through text.
	
	\noindent\textbf{Human evaluation} We evaluated StarGAN* and DWC-GAN by a user study where more than 20 people where asked to judge the Accuracy and Realism of generated results. On average, 82.0\% of the images generated by DWC-GAN were judged as correct, while StarGAN* generated only 25.6\% of corrected results. 
	Regarding Realism, DWC-GAN had an average score of 2.25, while StarGAN* had 1.17. Specifically, in 50.50\% of the people said DWC-GAN were perfect, recognizable and without defects while only 2.82\% of the people said the same for StarGAN*. StarGAN* results are indeed often judged ``neutral, recognizable but with obvious defects" (52.91\% of times). StackGAN++ and TAGAN do not achieve accuracy and realism higher than zero because they do not generate recognizable images. We refer to the Supplementary Material for additional details.
	
	These results show that DWC-GAN generates correct images, which can be mistaken with the real images at higher rates than competitors.

	\begin{table}[ht]
		\setlength{\tabcolsep}{5pt}
		\small
		\centering
		\begin{tabular}{@{}lrrrr@{}}
			\toprule
			\textbf{Method} & {\bf IS}$\uparrow$ & {\bf FID}$\downarrow$ & {\bf LPIPS}$\uparrow$ &  \textbf{Params~$\downarrow$}\\
			\midrule
			StackGAN++~\cite{zhang2018stackgan++} & 1.029 & 278.60 & .028$\pm$.009 &  50.30M\\
			TAGAN~\cite{nam2018text}  & 4.451 & 50.51 & .060$\pm$.024 &  44.20M \\
			ManiGAN~\cite{li2019manigan}  & 4.136 & 11.74 &  .001$\pm$.000 & 163.34M \\
			StarGAN*~\cite{choi2018stargan} & 4.343 & 109.89 & .000$\pm$.000 & 54.10M \\
			DWC-GAN (proposed) & \textbf{4.599} & \textbf{2.96} & \textbf{.081$\pm$.001} &  \textbf{33.53M} \\ 
			\bottomrule
		\end{tabular}
		\caption{Quantitative evaluation on the CUB dataset.}
		\label{tab:QuantitativeResults_cub} 
		\vspace{-5mm}
	\end{table} 
	
	\noindent\textbf{Progressive Manipulation}
	A few state of the art approaches for text-to-image generation and image manipulation are expressly designed for conversational systems~\cite{cheng2018sequential, el2018keep, el2019tell, li2019storygan, mogadala2019trends}.
	In Figure~\ref{fig:progressive} we show that we can repeatedly apply our method to generated images to have a manipulation in multiples steps. This result highlights the consistency of our generated results, which can be used as input of our model, and that DWC-GAN might be applied in an interactive environment.
	
\begin{figure}[ht]	
		\renewcommand{\tabcolsep}{1pt}
		\renewcommand{\arraystretch}{0.8}
		\newcommand{\sizea}{0.16\columnwidth}
		\footnotesize
		\begin{tabular}{c|cccccccc}
			Input Image & \multicolumn{2}{l}{\specialcell{``\textit{change the blond hair}\\\textit{to brown}"}} & $\longleftrightarrow$ & \multicolumn{2}{r}{\specialcellr{``\textit{blond hair,}\\ \textit{make her older}"}}\\
			\includegraphics[width=\sizea]{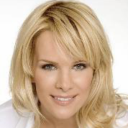} &
			\includegraphics[width=\sizea]{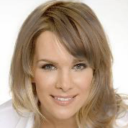} &
			\includegraphics[width=\sizea]{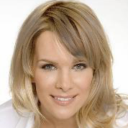} &
			\includegraphics[width=\sizea]{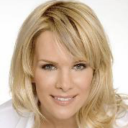} &
			\includegraphics[width=\sizea]{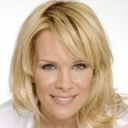} &
			\includegraphics[width=\sizea]{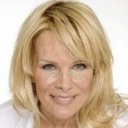} \\
			& \multicolumn{2}{l}{``\textit{grey breast}"} & $\longleftrightarrow$ & \multicolumn{2}{r}{``\textit{don't change anything}"}\\
			\includegraphics[width=\sizea]{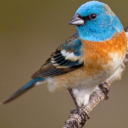} & 
			\includegraphics[width=\sizea]{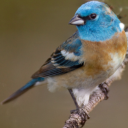} &
			\includegraphics[width=\sizea]{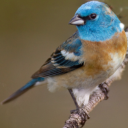} &
			\includegraphics[width=\sizea]{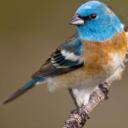} & 
			\includegraphics[width=\sizea]{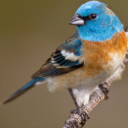} &
			\includegraphics[width=\sizea]{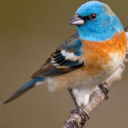}\\
		\end{tabular}
		\vspace{-5mm}
		\caption{Domain interpolation given an input image.}
		\label{Fig:interpolation}
	\end{figure}

	\noindent\textbf{Domain Interpolation} Our method generates interpolated images from two sentences. Figure~\ref{Fig:interpolation} shows that by manipulating an image in two different ways, it is possible to generate the intermediate steps that go from one textual manipulation to the other. 
	
	\begin{figure}[!ht]
		\centering
		\includegraphics[width=\linewidth]{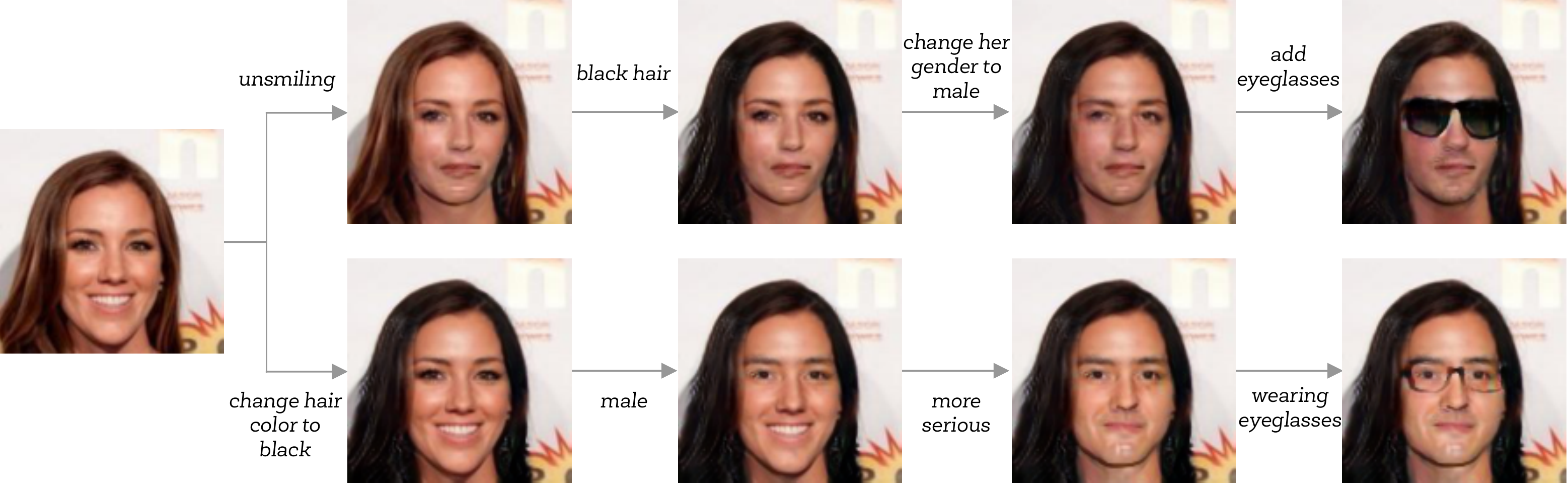}
		\vspace{-5mm}
		\caption{An example of progressive manipulation. Our method can used in an interactive environment.}
		\label{fig:progressive}
	\end{figure}
	
	\noindent \textbf{User Study on Textual Commands}
	Our training corpus is automatically generated, thus it might suffer limited generalizability. We conduct an experiment to verify these possible issues by asking 15 different people to describe in natural language how to edit an image A to make it similar to another image B. We asked people to do this experiment for 15 randomly chosen images. 
	The resulting collected textual commands and their corresponding images are fed into our model to evaluate the results. 
	We found that our model can be generalized to real human-written descriptions well. Details are presented in the Supplementary Material.
	
\begin{figure}[!ht]
		\centering
		\includegraphics[width=0.8\linewidth]{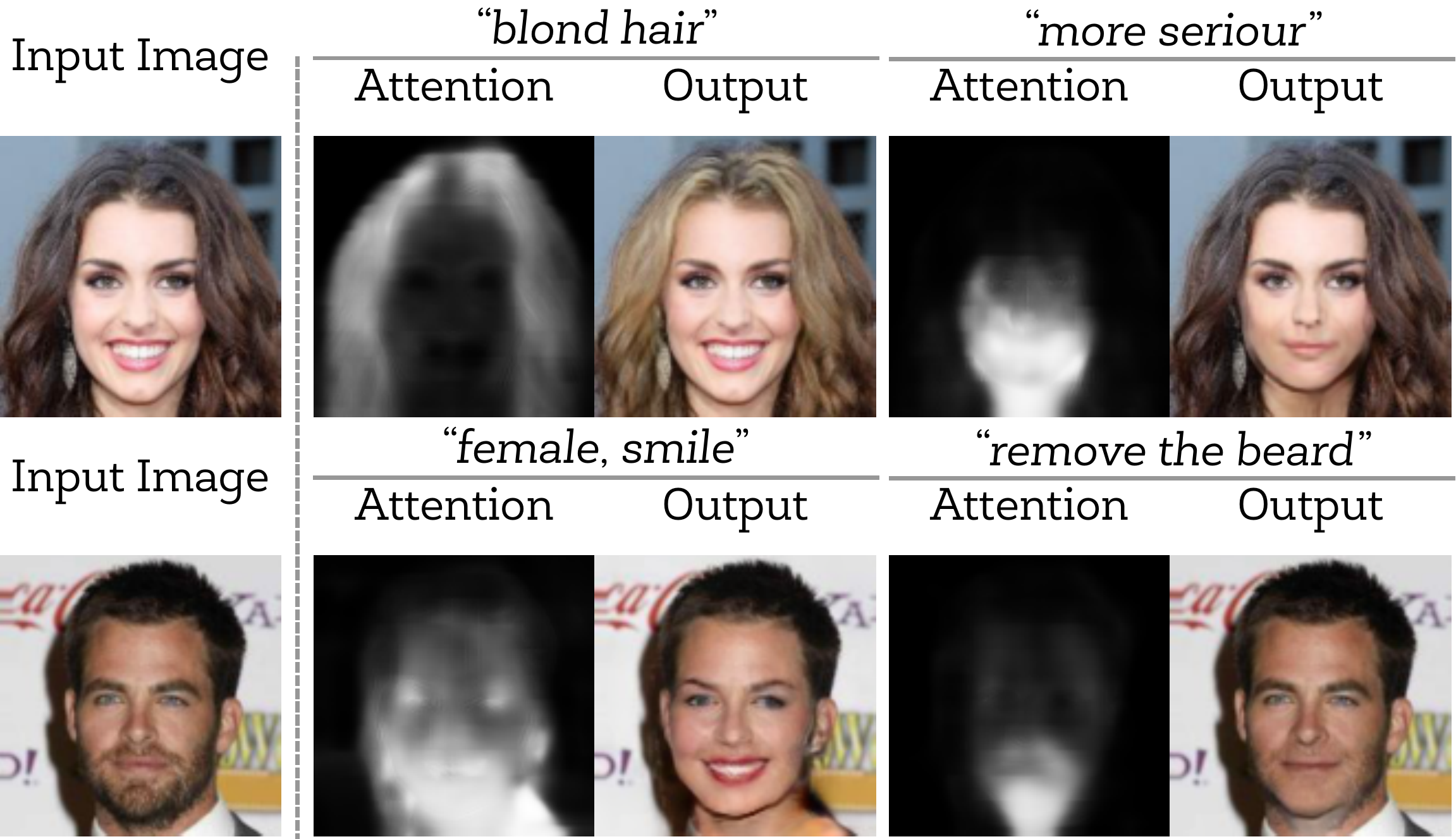}
		\vspace{-3mm}
		\caption{Unsupervised learned attention in DWC-GAN.}
		\label{fig:attention}
	\end{figure}

	\noindent\textbf{Attention Visualization} We visualize the unsupervised attention mask in Figure~\ref{fig:attention}. It shows the localization area of different manipulations, which indeed helps to focus on the attribute-related pixels and preserves the background pixels.
	
	\noindent\textbf{Ablation Study}
	We here investigate the effect of removing some components from our architecture. Results are showed in Table~\ref{tab:ablation_digits_details}.
	First, we remove the \emph{Cycle consistency}, which is widely employed by the literature of image-to-image translation~\cite{huang2018multimodal, zhu2017unpaired,  zhu2017toward}. 
	We observe that both the image quality and diversity significantly performs worse without the cycle consistency, which indicates that the Cycle consistency loss indeed constraints the network on the translation. 
	Then, when we remove the \emph{Attribute reconstruction}, which is applied to learn the attribute extractor $E_a$. Here, we use an attribute classifier to measure the F1 score. It shows that the constraint on the attribute latent variable slightly decreases the classification performance (see the Supplementary Material for details on the attribute classifier). However, both image quality and diversity are slightly improved.
	We also evaluate the contribution of pre-trained word embeddings by using our model without them. We observe that the use of pre-trained word embeddings improves the quality of generated images and converge the training faster. 
	We observe that diversity sensitive loss indeed increases the diversities during the sampling, while slightly reduces the performances on FID and IS. In additional, $\mathcal{L}_{cyc}, \mathcal{L}_{rec\_a}$ and using pre-trained embeddings can also have effects on the diversities.
	We investigate whether the contribution of our model relies only on the use of the GMM to model the attributes. Thus, we constrain down our model to a StarGAN*-like model by setting the covariance matrix of the GMM to zero ($\forall k\ \pmb{\Sigma}^k=0$). As expected, we observe that our model behaves similarly but with no diversity (LPIPS: .000$\pm$.000). Moreover, this proves that our model outperforms StarGAN* even in this case (see Table~\ref{tab:QuantitativeResults_ce} for a comparison).
	
	\begin{table}[!ht]
		\small
		\centering
		\begin{tabular}{@{}lrrrr@{}}
			\toprule
			\textbf{Model} & \textbf{IS$\uparrow$} & \textbf{FID$\downarrow$} & \textbf{LPIPS$\uparrow$} & \textbf{F1 (\%)$\uparrow$}\\ 
			\midrule 
			DWC-GAN w/o $\mathcal{L}_{cyc}$ & 2.589 & 97.94 & .042$\pm$.001  & 96.32 \\ 
			DWC-GAN w/o embedd. & 2.782 & 73.90 & .033$\pm$.001 &  98.06 \\ 
			DWC-GAN w/o $\mathcal{L}_{rec\_a}$ & 2.961 & 32.35 & .072$\pm$.001 & 98.49 \\ 
			DWC-GAN $\forall k\ \pmb{\Sigma}^k=0$ & 2.872 & 33.17 & .000$\pm$.000 & \textbf{98.64} \\ 
			DWC-GAN w/o $\mathcal{L}_{ds}$ & \textbf{3.148} & \textbf{31.29} & .061$\pm$.001 & 94.69\\ 
			DWC-GAN & 3.069 & 32.14 & \textbf{.152$\pm$.003} & 94.80 \\ 
			\bottomrule
		\end{tabular}
		\caption{Ablation study performance on the CelebA dataset. 
		}
		\label{tab:ablation_digits_details}
		\vspace{-5mm}
	\end{table}
	
	\section{Conclusion}
	In this paper, we presented a novel method to manipulate visual attributes through text. Our model builds upon recent literature of image to image translation, using unpaired images and modeling visual attributes through a GMM. 
	Users interact with the model by feeding an image and a text that describes the desired alteration to the original image. 
	The model extracts from the input image the visual attributes, understand the text and the desired transformation, and sample multiple images having the content of the input image and the style of the desired output. 
	By sampling multiple synthesized images, our model deals with the inherent ambiguity of natural language and shows the user multiple transformations (e.g. multiple shadows and styles of blonde hair) from which she/he might select the best output.
	
	To the best of our knowledge, we are the first at exploiting automatically generated sentences and unpaired images to manipulate image attributes with textual commands. We showed that our method can model multiple attributes, and that can be applied in a wide range of situations such as interpolation and progressive manipulation. We foresee that our work will stimulate further research on attribute-based image manipulation and generation and will stimulate software integration in virtual assistants such as Alexa or image editors such as Photoshop.

	\bibliographystyle{ACM-Reference-Format}
	\bibliography{egbib}
	\clearpage

\appendix

\section{Implementation Details}
\label{suppl:implementation_details}

We built our model upon the recent state-of-the-art approaches, namely MUNIT~\cite{huang2018multimodal}, BicycleGAN~\cite{zhu2017toward}, StarGAN~\cite{choi2018stargan} and TAGAN~\cite{nam2018text}. We apply Instance Normalization (IN)~\cite{ulyanov2017improved}
to the content encoder $E_c$, while we use Adaptive Instance Normalization (AdaIN)~\cite{huang2017arbitrary} and Layer Normalization (LN)~\cite{ba2016layer} for the decoder $G$. We use a 2 layers LSTM~\cite{hochreiter1997long} with Dropout~\cite{srivastava2014dropout} rate 0.1 for the text encoder $E_t$. For the discriminator network, we use Leaky ReLU~\cite{xu2015empirical} with a negative slope of 0.2. The details of the architecture are shown in Table~\ref{tab:architecture}.
For the test of the two datasets, we use 2 scales for the discriminator, in which the resolutions of input images are 128$\times$128 and 64$\times$64, respectively.

\begin{table*}[!ht]
	\centering
	\begin{tabular}{@{}lll@{}}
	    \toprule
	    \textbf{Part} & \textbf{Input} $\rightarrow$ \textbf{Output Shape} & \textbf{Layer Information} \\ \midrule
		\multirow{7}{*}{$E_c$} & ($h$, $w$, 3) $\rightarrow$ ($h$, $w$, 64) & CONV-(N64, K7x7, S1, P3), IN, ReLU \\
		& ($h$, $w$, 64) $\rightarrow$ ($\frac{h}{2}$, $\frac{w}{2}$, 128) & CONV-(N128, K4x4, S2, P1), IN, ReLU \\
		& ($\frac{h}{2}$, $\frac{w}{2}$, 128) $\rightarrow$ ($\frac{h}{4}$, $\frac{w}{4}$, 256) & CONV-(N256, K4x4, S2, P1), IN, ReLU \\ 
		& ($\frac{h}{4}$, $\frac{w}{4}$, 256) $\rightarrow$ ($\frac{h}{4}$, $\frac{w}{4}$, 256) & Residual Block: CONV-(N256, K3x3, S1, P1), IN, ReLU \\ 
		& ($\frac{h}{4}$, $\frac{w}{4}$, 256) $\rightarrow$ ($\frac{h}{4}$, $\frac{w}{4}$, 256) & Residual Block: CONV-(N256, K3x3, S1, P1), IN, ReLU \\ 
		& ($\frac{h}{4}$, $\frac{w}{4}$, 256) $\rightarrow$ ($\frac{h}{4}$, $\frac{w}{4}$, 256) & Residual Block: CONV-(N256, K3x3, S1, P1), IN, ReLU \\ 
		& ($\frac{h}{4}$, $\frac{w}{4}$, 256) $\rightarrow$ ($\frac{h}{4}$, $\frac{w}{4}$, 256) & Residual Block: CONV-(N256, K3x3, S1, P1), IN, ReLU \\ \midrule
		\multirow{9}{*}{$E_a$} & ($h$, $w$, 3) $\rightarrow$ ($h$, $w$, 64) & CONV-(N64, K7x7, S1, P3), ReLU \\
		& ($h$, $w$, 64) $\rightarrow$ ($\frac{h}{2}$, $\frac{w}{2}$, 128) & CONV-(N128, K4x4, S2, P1), ReLU \\
		& ($\frac{h}{2}$, $\frac{w}{2}$, 128) $\rightarrow$ ($\frac{h}{4}$, $\frac{w}{4}$, 256) & CONV-(N256, K4x4, S2, P1), ReLU \\ 
		& ($\frac{h}{4}$, $\frac{w}{4}$, 256) $\rightarrow$ ($\frac{h}{8}$, $\frac{w}{8}$, 256) & CONV-(N256, K4x4, S2, P1), ReLU \\
		& ($\frac{h}{8}$, $\frac{w}{8}$, 256) $\rightarrow$ ($\frac{h}{16}$, $\frac{w}{16}$, 256) & CONV-(N256, K4x4, S2, P1), ReLU \\
		& ($\frac{h}{16}$, $\frac{w}{16}$, 256) $\rightarrow$ ($\frac{h}{32}$, $\frac{w}{32}$, 256) & CONV-(N256, K4x4, S2, P1), ReLU \\
		& ($\frac{h}{32}$, $\frac{w}{32}$, 256) $\rightarrow$ (1, 1, 256) & GAP \\ \cmidrule{2-3}
		& (256,) $\rightarrow$ ($Z$,) & FC-(N$Z$) \\
		& (256,) $\rightarrow$ ($Z$,) & FC-(N$Z$) \\
		\midrule
		\multirow{2}{*}{$E_t$} & (300(+$Z$), T) $\rightarrow$ (2400,) & LSTM(W300, H300, L2, BiD, Dropout) \\ 
		& (2400,) $\rightarrow$ ($Z$,) & FC-(N$Z$)\\ \midrule
		\multirow{7}{*}{$G$} & ($\frac{h}{4}$, $\frac{w}{4}$, 256) $\rightarrow$ ($\frac{h}{4}$, $\frac{w}{4}$, 256) & Residual Block: CONV-(N256, K3x3, S1, P1), AdaIN, ReLU \\
		& ($\frac{h}{4}$, $\frac{w}{4}$, 256) $\rightarrow$ ($\frac{h}{4}$, $\frac{w}{4}$, 256) & Residual Block: CONV-(N256, K3x3, S1, P1), AdaIN, ReLU \\
		& ($\frac{h}{4}$, $\frac{w}{4}$, 256) $\rightarrow$ ($\frac{h}{4}$, $\frac{w}{4}$, 256) & Residual Block: CONV-(N256, K3x3, S1, P1), AdaIN, ReLU \\
		& ($\frac{h}{4}$, $\frac{w}{4}$, 256) $\rightarrow$ ($\frac{h}{4}$, $\frac{w}{4}$, 256) & Residual Block: CONV-(N256, K3x3, S1, P1), AdaIN, ReLU \\
		& ($\frac{h}{4}$, $\frac{w}{4}$, 256) $\rightarrow$ ($\frac{h}{2}$, $\frac{w}{2}$, 128) & UPCONV-(N128, K5x5, S1, P2), LN, ReLU \\ 
		 & ($\frac{h}{2}$, $\frac{w}{2}$, 128) $\rightarrow$ ($h$, $w$, 64) & UPCONV-(N64, K5x5, S1, P2), LN, ReLU \\ 
		 & (${h}$, ${w}$, 64) $\rightarrow$ (${h}$, ${w}$, 3) & CONV-(N3, K7x7, S1, P3), Tanh \\ \midrule
		 \multirow{7}{*}{$D$} & ($h$, $w$, 3) $\rightarrow$ ($\frac{h}{2}$, $\frac{w}{2}$, 64) & CONV-(N64, K4x4, S2, P1), Leaky ReLU \\
		 & ($\frac{h}{2}$, $\frac{w}{2}$, 64) $\rightarrow$ ($\frac{h}{4}$, $\frac{w}{4}$, 128) & CONV-(N128, K4x4, S2, P1), Leaky ReLU \\
		 & ($\frac{h}{4}$, $\frac{w}{4}$, 128) $\rightarrow$ ($\frac{h}{8}$, $\frac{w}{8}$, 256) & CONV-(N256, K4x4, S2, P1), Leaky ReLU \\
		 & ($\frac{h}{8}$, $\frac{w}{8}$, 256) $\rightarrow$ ($\frac{h}{16}$, $\frac{w}{16}$, 512) & CONV-(N512, K4x4, S2, P1), Leaky ReLU \\
		 & ($\frac{h}{16}$, $\frac{w}{16}$, 512) $\rightarrow$ ($\frac{h}{32}$, $\frac{w}{32}$, 512) & CONV-(N512, K4x4, S2, P1), Leaky ReLU \\\cmidrule{2-3}
		 & ($\frac{h}{32}$, $\frac{w}{32}$, 512) $\rightarrow$ ($\frac{h}{32}$, $\frac{w}{32}$, 1) & CONV-(N1, K1x1, S1, P0) \\ 
		 & ($\frac{h}{32}$, $\frac{w}{32}$, 512) $\rightarrow$ (1, 1, $n$) & CONV-(N$n$, K$\frac{h}{8}$x$\frac{w}{8}$, S1, P0) \\ 
		\bottomrule
	\end{tabular}
	\vspace{1mm}
	\caption{Network architecture. We use the following notation: $Z$: the dimension of attribute vector, $n$: the number of the domains, N: the number of output channels, K: kernel size, S: stride size, P: padding size, CONV: a convolutional layer, GAP: a global average pooling layer, UPCONV: a 2$\times$ bilinear upsampling layer followed by a convolutional layer, FC: fully connected layer, T: the number of the words in the text sentence, W: the dimension of word embedding, H: hidden size of the LSTM layer, BiD: using bidirectional encoding, L: the number of LSTM layers.}
	\label{tab:architecture}
\end{table*}

We use the Adam optimizer~\cite{kingma2014adam} with $\beta_1$ = 0.5, $\beta_2$ = 0.999, and an initial learning rate of 0.0001. The learning rate is decreased by half every 2e5 iterations. We set maximal iterations 1e6 for CelebA and 3e5 for CUB dataset. In all experiments, we use a batch size of 1. And we set the loss weights to $\lambda_{rec\_s}$ = 10, $\lambda_{cyc}$ = 10, and $\lambda_{KL}$ = 0.1. 
Random mirroring is applied during training.

In the experiments, we use a simplified version of the GMM, which satisfies the following properties:
\begin{itemize}
    \item The mean vectors are placed on the vertices of $Z$-dimensional regular simplex, so that the mean vectors are equidistant (and disentangled). Here $Z = d*K$ from the main paper.
    \item The covariance matrices are diagonal, with the same on all the components. In other words, each Gaussian component is \textit{spherical}, formally: $\pmb{\Sigma}_k = \pmb{\sigma}_k\mathbf{I}$, where $\mathbf{I}$ is the identity matrix.
\end{itemize}
Changing the parameter $\pmb{\sigma}_k$ to a number $\lambda \pmb{\sigma}_k$ does not change the model as in normal VAEs~\cite{doersch2016tutorial}. However, when $\lambda$ is bigger than the distance between means in the simplex, this poses issues on the assumption of independent and disentangled latent factors.

Regarding the baseline models, we modify the baseline models based on their corresponding released codes, including StackGAN++\footnote[1]{https://github.com/hanzhanggit/StackGAN-v2}, StarGAN\footnote[2]{https://github.com/yunjey/stargan},  TAGAN\footnote[3]{https://github.com/woozzu/tagan} and ManiGAN\footnote[4]{https://github.com/mrlibw/ManiGAN}. To train StarGAN on this task, we add a module $E_t$ same as ours and learn to predict target attributes guided by binary cross-entropy loss. For fairness, all models take use the same corpus and images during the training. 
To test the training speed in Table 1 of the main paper we set the batch size 1 and use a single GTX TITAN Xp with 12GB Memory for all models.

We release the source code of our model and the StarGAN* implementation at \url{https://xxx}~\footnote[5]{Temporary attached in the submission as supplementary file to keep the anonymity of the authors. It will be stored in a permanent link upon acceptance.}

\section{Attribute classifier}
We evaluate the consistency of attributes labels through a model that is trained to classify faces on the four tested combinations of attributes. We finetune an ResNet-50~\cite{he2016deep} model trained on ImageNet~\cite{deng2009imagenet} by using the $L^2$-SP regularization~\cite{li2018} on all the layers, with $\alpha = 0.01$ and $\beta = 0.01$. The list of training examples is the same as the main model, thus keeping the hold-out dataset as unseen. From the training list we carved out $10\%$ of the data for the validation and $20\%$ for testing.

We use the Adam optimizer~\cite{kingma2014adam} with $\beta_1$ = 0.5, $\beta_2$ = 0.999, and an initial learning rate of 0.001. The learning rate is divided by 10 whenever the training loss does not decrease for two epochs. The training is early stopped when the validation loss does not decrease after 10 epochs.

We apply a central crop and random rotation until at most 10 degrees data augmentation techniques during training. The images are then resized to $299 \times 299$ pixels.
The accuracy is $0.91$ and $0.91$ in the training and test set respectively.

We use this classifier to evaluate the accuracy of our transformations through the Micro-F1 (the F1 score that weights equally all test samples) score between the desired attributes and the generated ones.

\section{Automatic Text Protocol} 
\label{suppl:automatic_text}
We propose an automatic text protocol to collect the description text, which takes use of the attribute annotations only. Hence, we launch our experiments based on two multi-label recognition or classification~\cite{liu2015deep,wah2011caltech} datasets. In CelebA~\cite{liu2015deep} dataset, we select 8 attributes from annotated attributes \{  ``\textit{Black\_Hair}", ``\textit{Blond\_Hair}", ``\textit{Brown\_Hair}", ``\textit{Male}", ``\textit{Smiling}", ``\textit{Young}", ``\textit{Eyeglasses}", ``\textit{No\_Beard}"\}, which including both the style (e.g. hair color) and content (e.g. gender, age, beard and eyeglasses).  In CUB~\cite{wah2011caltech} dataset, we select three body parts of the birds \{``\textit{wing}", ``\textit{crown}", ``\textit{breast}"\} and each part can be consist of a set of colors \{``\textit{grey}", ``\textit{brown}", ``\textit{red}", ``\textit{yellow}", ``\textit{white}", ``\textit{buff}"\}. 

\begin{figure*}[!ht]
    \centering
    \includegraphics[width=0.7\textwidth]{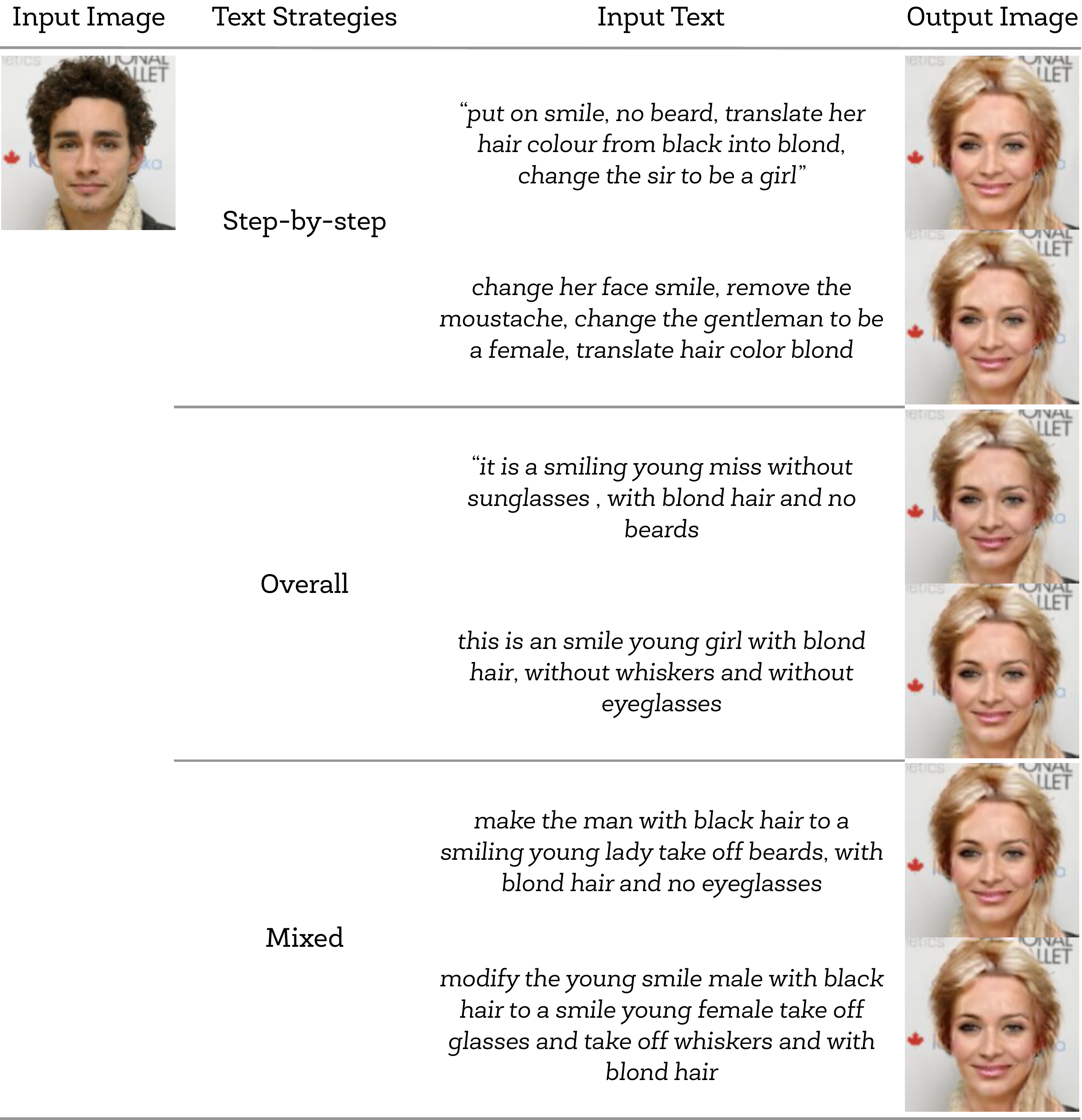}
    \caption{Examples on automatic text strategies. It shows a consistent performance with various textual commands for the same manipulation.}
    \label{fig:appendix-text}
\end{figure*}

To automatically collect manipulation text, we provide some common templates and phrases to collect the text, such as ``\textit{change ... to ...}", ``\textit{with ... hair}", ``\textit{make the face/bird ...}", ``\textit{a man/woman/bird with/has ...}". We randomly permute the description sequence of the attributes and replace some words with synonymy (e.g. ``\textit{change}", ``\textit{translate}", ``\textit{modify}" and ``\textit{make}") to improve the generalization. 
Similarly, we randomly select the synonymy named entities during generate text (e.g., ``\textit{man}", ``\textit{male}", ``\textit{boy}", ``\textit{gentleman}" and ``\textit{sir}" for the attribute \textit{Male}). As shown in Figure~\ref{fig:appendix-text}, it indicates that our model generates consistent results with various textual commands for the same manipulation. We also test the sensitivity of the strategy, as the choice of the strategy might influence the results of the model. Table~\ref{tab:sensitivity_strategy} shows that our model is robust to the choice of the textual strategy. 

\begin{table}[!ht]
	\centering
	\begin{tabular}{@{}lr r r@{}}
	\toprule
		\multirow{2}{*}{\textbf{Strategy Name}} &
		\multicolumn{3}{c}{\textbf{Metrics}}  \\
		\cmidrule{2-4}
		& IS$\uparrow$ & FID$\downarrow$ & F1$\uparrow$ \\
		\midrule
		Step-by-step & 3.070 & 32.25 & 94.14 \\
		Overall & 3.081 & 32.07 & 94.78\\
		Mixed & 3.078 & 32.16 & 94.80 \\ 
		Random & 3.069 & 32.14 & 94.80  \\
		\bottomrule
	\end{tabular}
	\vspace{1mm}
	\caption{Quantitative comparison on the different automatic text strategies on the CelebA dataset.}
	\label{tab:sensitivity_strategy}
\end{table}

\begin{figure}[ht]
    \centering
    \includegraphics[width=\columnwidth]{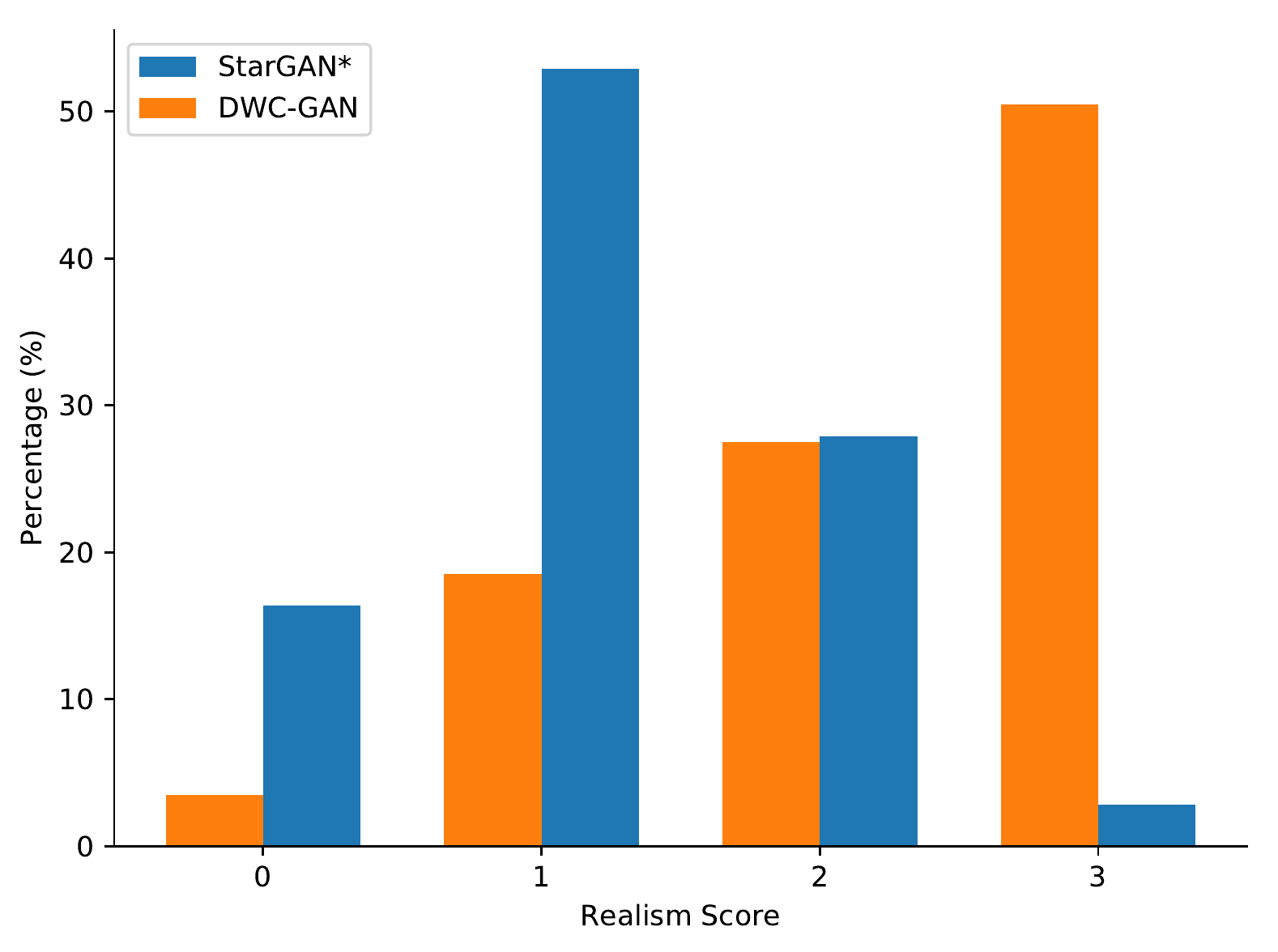}
    \caption{Distributions on the Realism human evaluation.}
    \label{fig:human_evaluation}
\end{figure}

\section{User Study}
\begin{figure*}[!ht]
    \centering
    \includegraphics[width=0.64\textwidth]{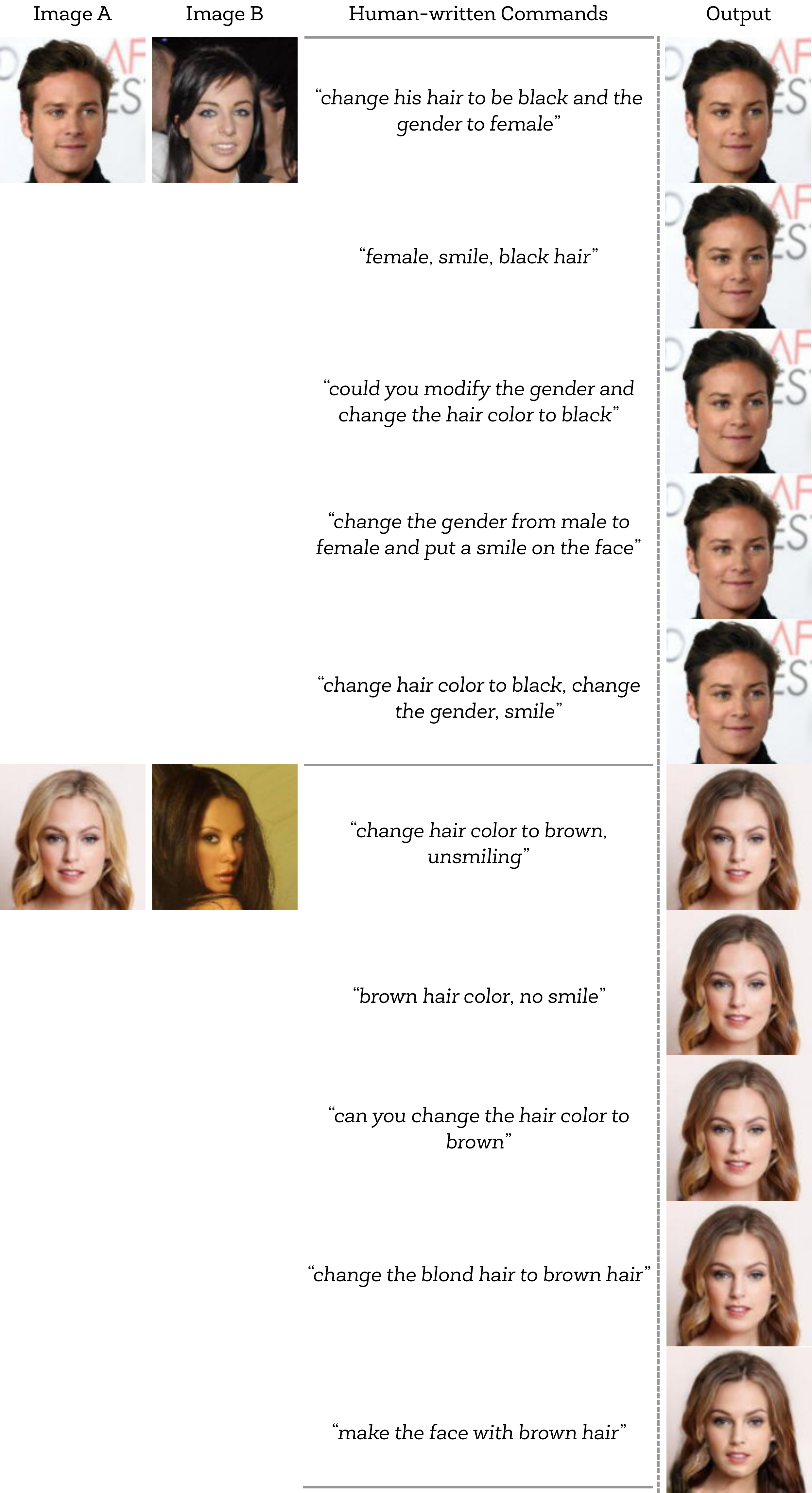}
    \caption{Examples on asking users to describe how to change image A to make it similar to image B.}
    \label{fig:appendix-user-text1}
\end{figure*}

\begin{figure*}[!ht]
    \centering
    \includegraphics[width=0.64\textwidth]{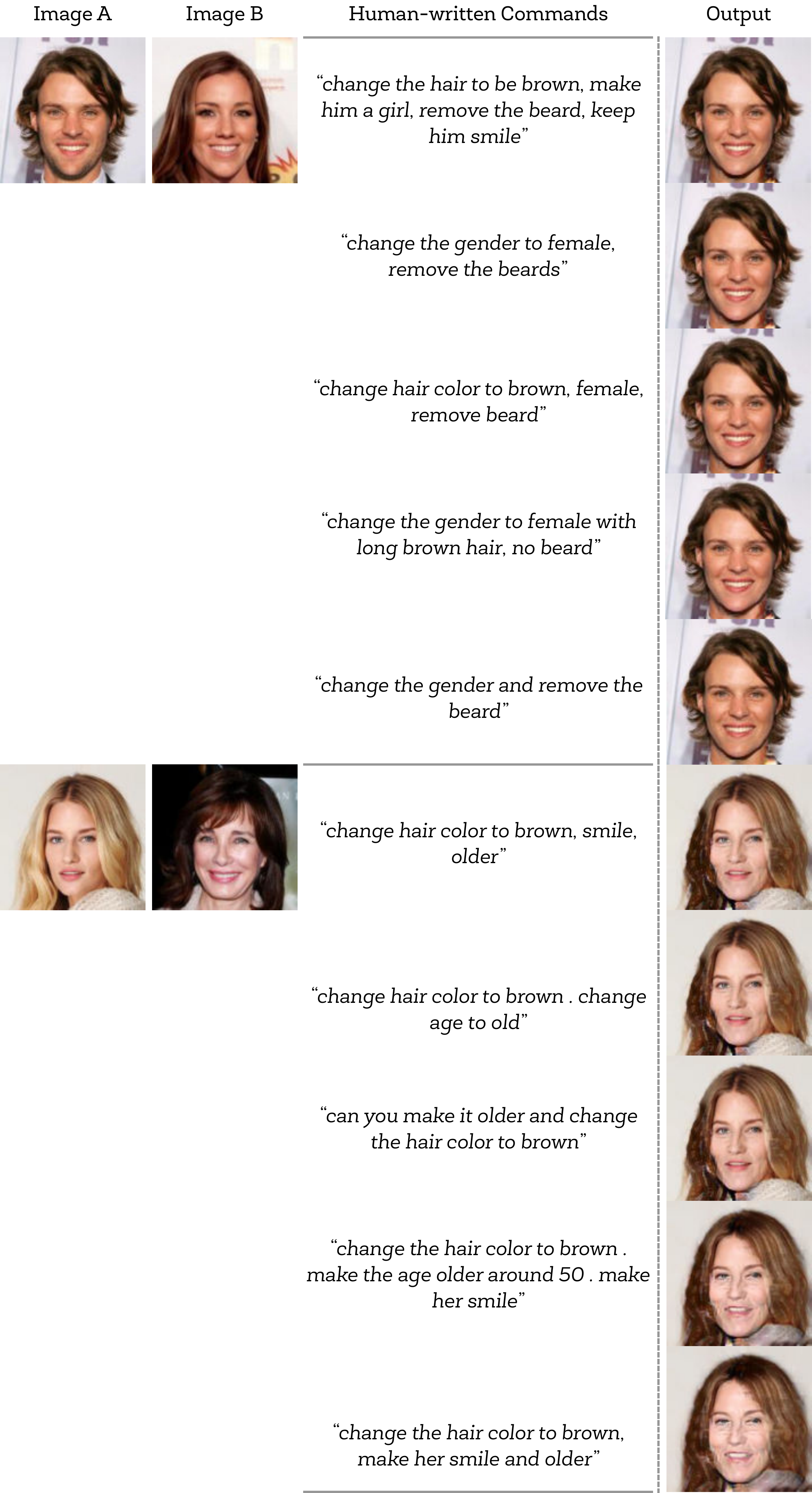}
    \caption{Examples on asking users to describe how to change image A to make it similar to image B.}
    \label{fig:appendix-user-text2}
\end{figure*}

We ask different people with various backgrounds to describe how to change the ``image A" to be similar to ``image B", as shown in Figure~\ref{fig:appendix-user-text1} and Figure~\ref{fig:appendix-user-text2}. In Figure~\ref{fig:human_evaluation} we show the result of the human evaluation on the realism.

\section{Domain Interpolation}
\begin{figure*}[ht]
    \centering
    \includegraphics[width=0.9\textwidth]{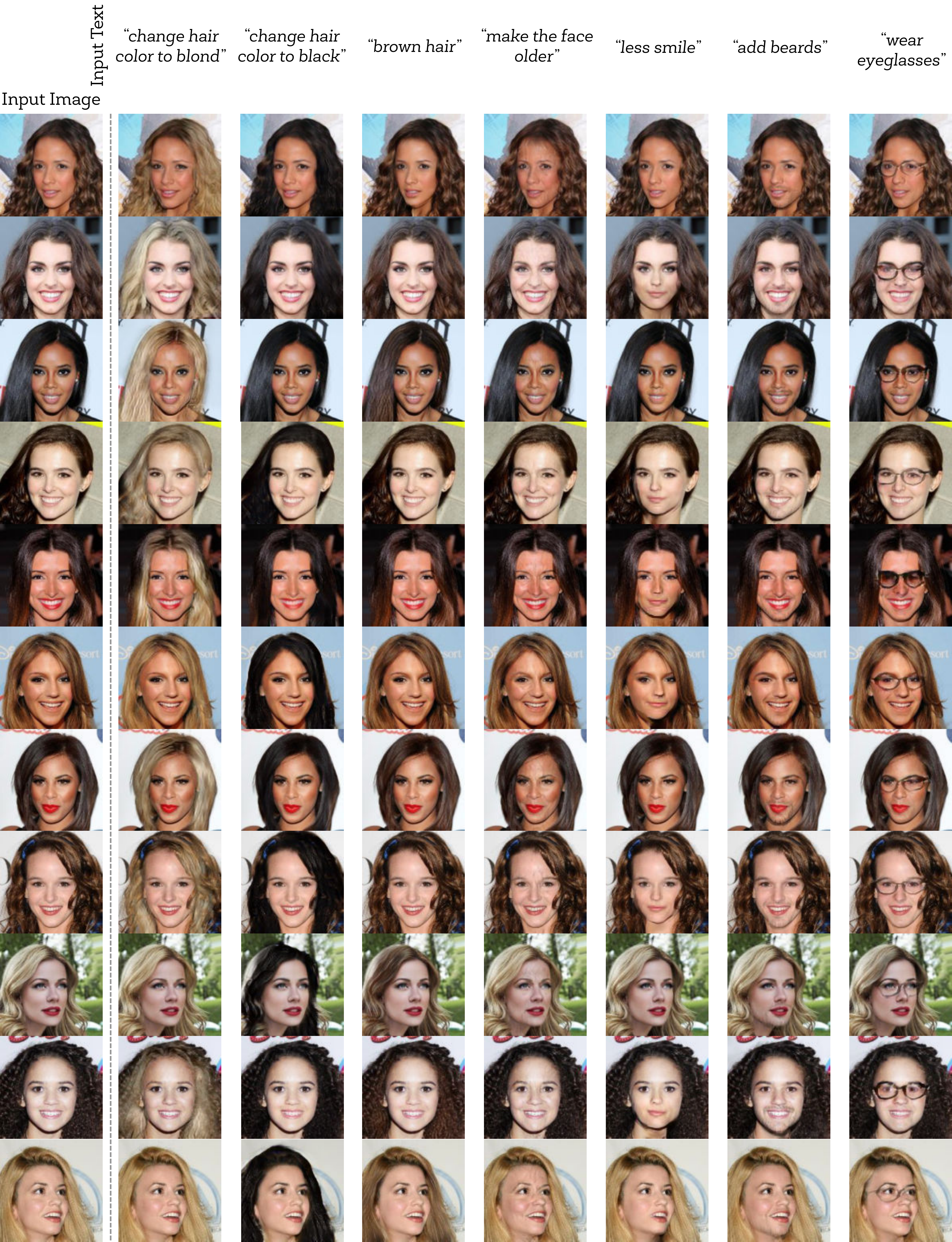}
    \caption{Examples of translations on CelebA dataset.}
    \label{fig:domain_translation_celeba1}
\end{figure*}

\begin{figure*}[ht]
    \centering
    \includegraphics[width=0.9\textwidth]{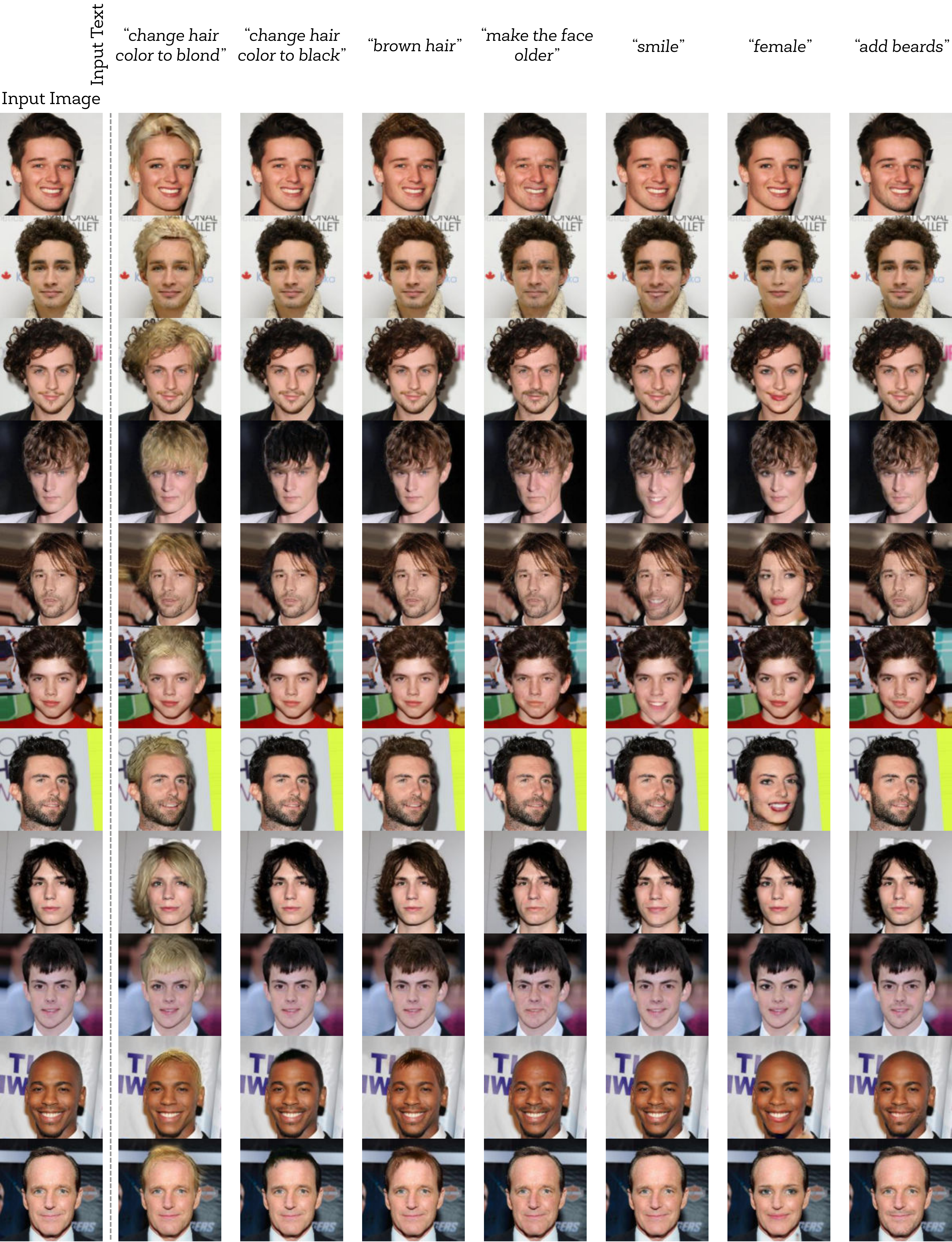}
    \caption{Examples of translations on CelebA dataset.}
    \label{fig:domain_translation_celeba2}
\end{figure*}

\begin{figure*}[ht]	
	\renewcommand{\tabcolsep}{1pt}
	\renewcommand{\arraystretch}{0.8}
	\newcommand{\sizea}{0.10\textwidth}
	\centering
	\footnotesize
	\begin{tabular}{c|cccccccc}
	    Input image 
		& \multicolumn{2}{l}{\textit{younger}}  & \multicolumn{2}{c}{$\longleftrightarrow$} &  \multicolumn{2}{r}{\textit{make the face more serious}}\\
		\includegraphics[width=\sizea]{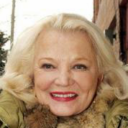} &
		\includegraphics[width=\sizea]{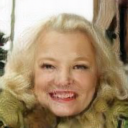}&
		\includegraphics[width=\sizea]{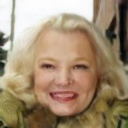} &
		\includegraphics[width=\sizea]{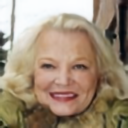} &
		\includegraphics[width=\sizea]{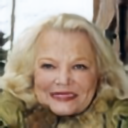} &
		\includegraphics[width=\sizea]{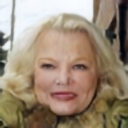} &
		\includegraphics[width=\sizea]{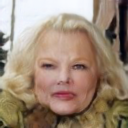}\\
		& \multicolumn{2}{l}{\textit{do not change anything}} & \multicolumn{2}{c}{$\longleftrightarrow$} & \multicolumn{2}{r}{\textit{no beards, more attractive}}\\
		\includegraphics[width=\sizea]{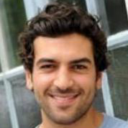} & 
		\includegraphics[width=\sizea]{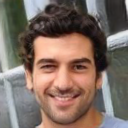} &
		\includegraphics[width=\sizea]{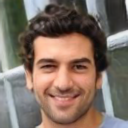} &
		\includegraphics[width=\sizea]{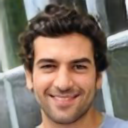} &
		\includegraphics[width=\sizea]{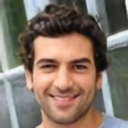} &
		\includegraphics[width=\sizea]{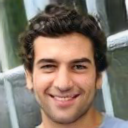} &
		\includegraphics[width=\sizea]{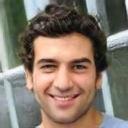}\\
		& \multicolumn{2}{l}{\textit{smiling}} & \multicolumn{2}{c}{$\longleftrightarrow$} & \multicolumn{2}{r}{\textit{increase his age}}\\
		\includegraphics[width=\sizea]{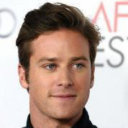} & 
		\includegraphics[width=\sizea]{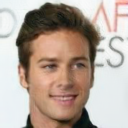} &
		\includegraphics[width=\sizea]{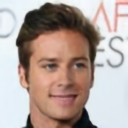} &
		\includegraphics[width=\sizea]{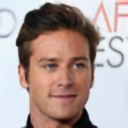} &
		\includegraphics[width=\sizea]{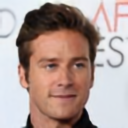} &
		\includegraphics[width=\sizea]{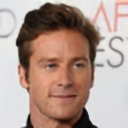} &
		\includegraphics[width=\sizea]{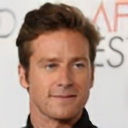}\\
	\end{tabular}
	\caption{Examples of domain interpolation given an input image.}
	\label{suppl:Fig:interpolation}
\end{figure*}

We show more results in Figure~\ref{fig:domain_translation_celeba1} and Figure~\ref{fig:domain_translation_celeba2} on various face images. 
Figure~\ref{suppl:Fig:interpolation} show examples of domain interpolation. 

\section{Domain Sampling}
\begin{figure*}[ht]
    \centering
    \includegraphics[width=0.62\textwidth]{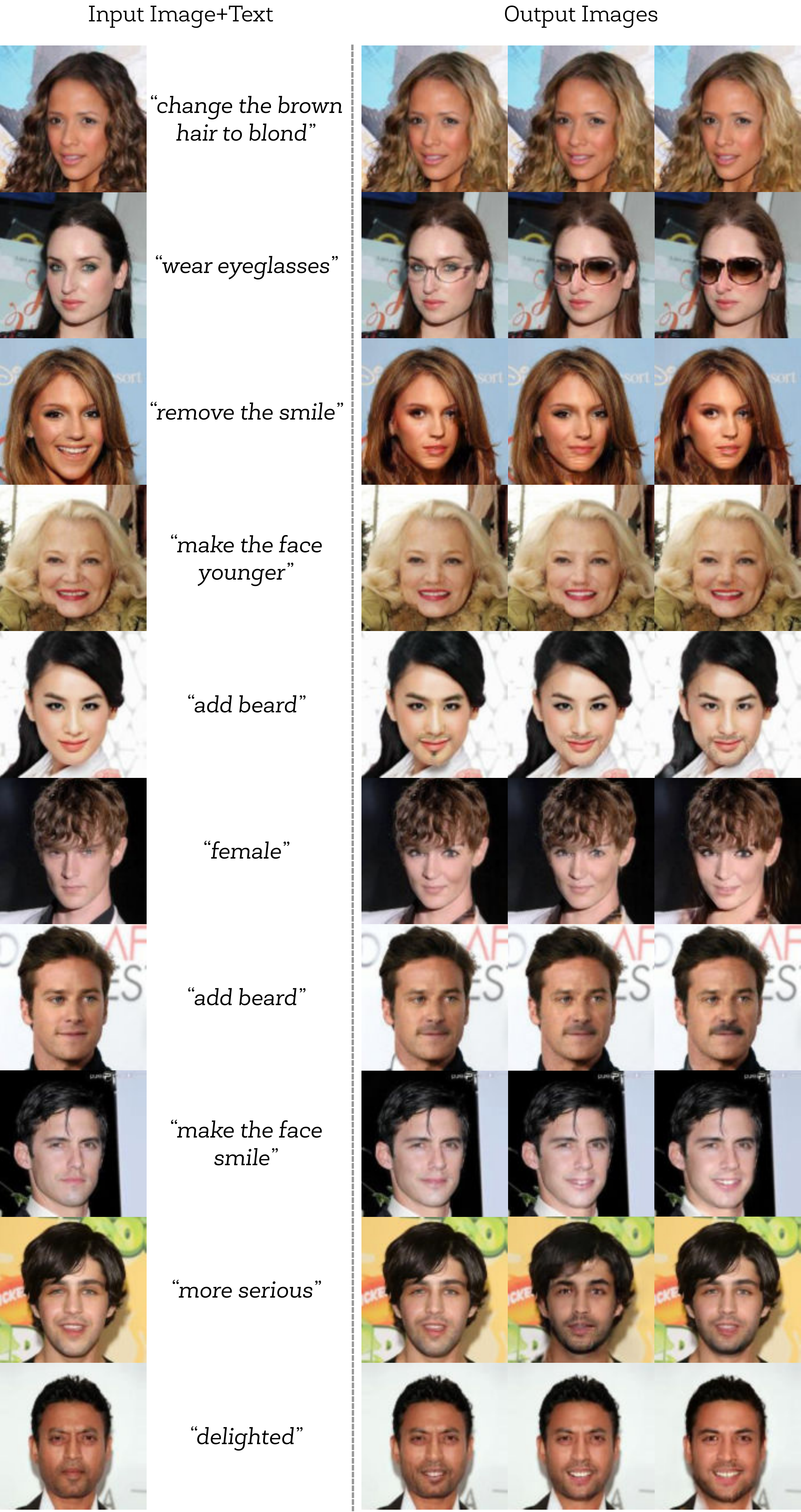}
    \caption{Examples of domain sampling on CelebA dataset.}
    \label{fig:domain_sampling}
\end{figure*}

As shown in Figure~\ref{fig:domain_sampling}, we apply the domain sampling to obtain diverse results that are consistent with the same textual commands.

\begin{figure*}[!ht]
	\renewcommand{\tabcolsep}{2pt}
	\renewcommand{\arraystretch}{0.8}
    \centering
    \newcommand{\sizea}{0.10\textwidth}
    {\small
    \begin{tabular}{c|ccccc}
	 Input & \multicolumn{5}{c}{``\textit{change the hair color to black and blond}"} \\
		\includegraphics[width=\sizea]{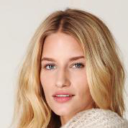} &
		\includegraphics[width=\sizea]{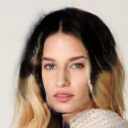} &
		\includegraphics[width=\sizea]{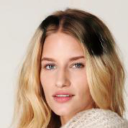} &
		\includegraphics[width=\sizea]{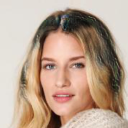} &
		\includegraphics[width=\sizea]{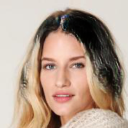} & \includegraphics[width=\sizea]{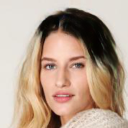} \\
        & \multicolumn{5}{c}{``\textit{a smiling attractive old lady with black and blond hair}"} \\
		\includegraphics[width=\sizea]{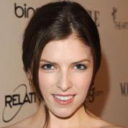} &
		\includegraphics[width=\sizea]{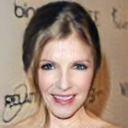} &
		\includegraphics[width=\sizea]{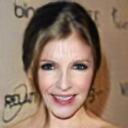} &
		\includegraphics[width=\sizea]{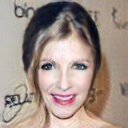} &
		\includegraphics[width=\sizea]{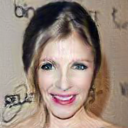} & \includegraphics[width=\sizea]{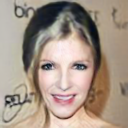} \\
	\end{tabular}}
    \caption{Generated images in previously unseen combinations of attributes.}
    \label{Fig:unseen}
\end{figure*}

We also evaluate the ability of DWC-GAN to synthesize new images with attributes that are extremely scarce or non present in the training dataset, as shown in Figure~\ref{Fig:unseen}. To do so, we select three combinations of attributes consisting of less than two images in the CelebA dataset: $[1,1,0,0,1,1,0,1]$ (``\textit{Black\_Hair, Blond\_Hair, Female, Smiling, Young, No\_Eyeglasses, No\_Beard}") and $[1,1,0,0,1,0,0,1]$ (``\textit{Black\_Hair, Blond\_Hair, Female, Smiling, Old, No\_Eyeglasses, No\_Beard}"). We show that the model can effectively generate configurations in previously unseen domains if we describe them trough the textual condition. These generated images show that our model can effectively do few- or zero-shot generation.

\begin{figure*}[ht]	
	\renewcommand{\arraystretch}{4}
	\centering
	\footnotesize
	\begin{tabular}{c}
    \includegraphics[width=0.85\linewidth]{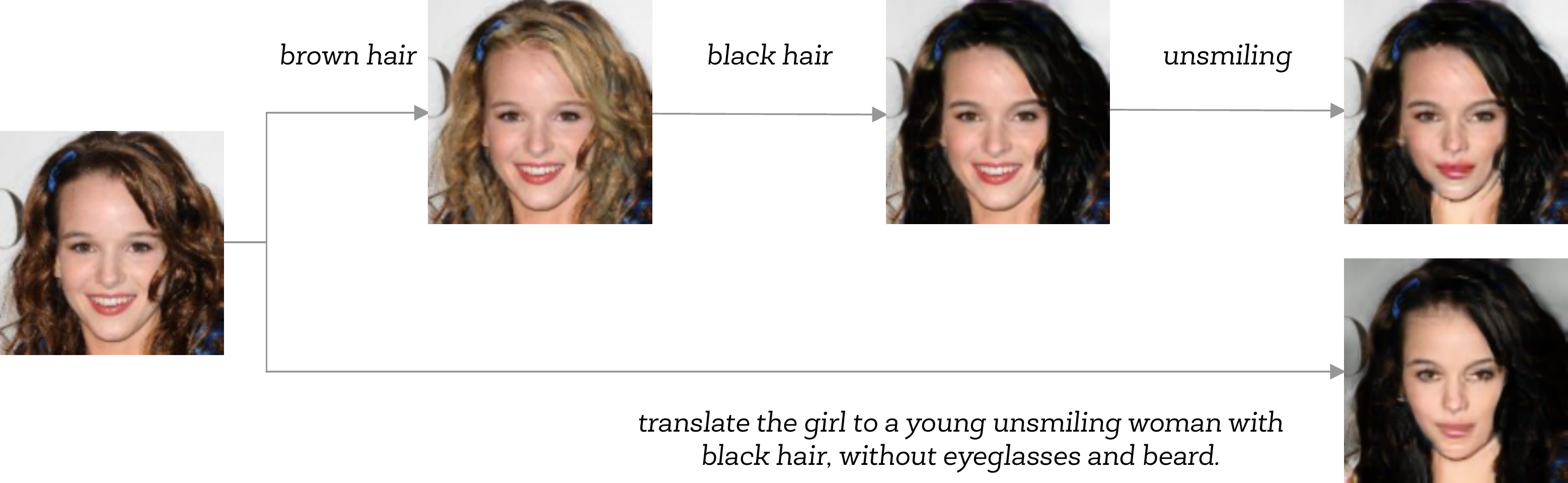} \\
    \includegraphics[width=0.85\linewidth]{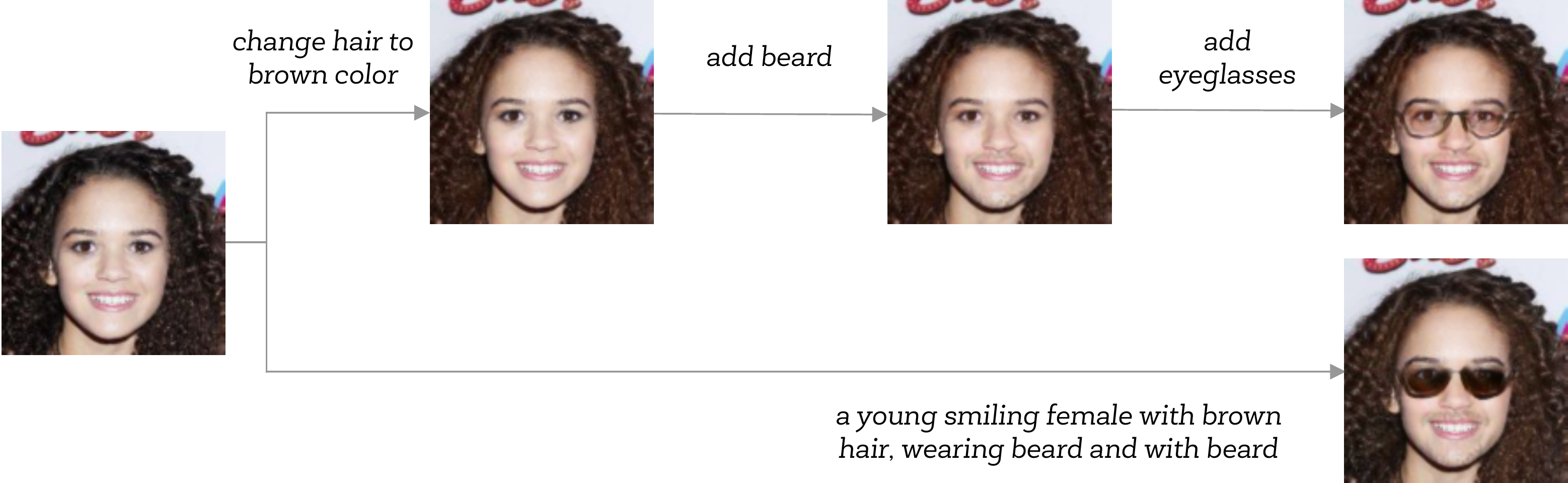} \\
    \includegraphics[width=0.85\linewidth]{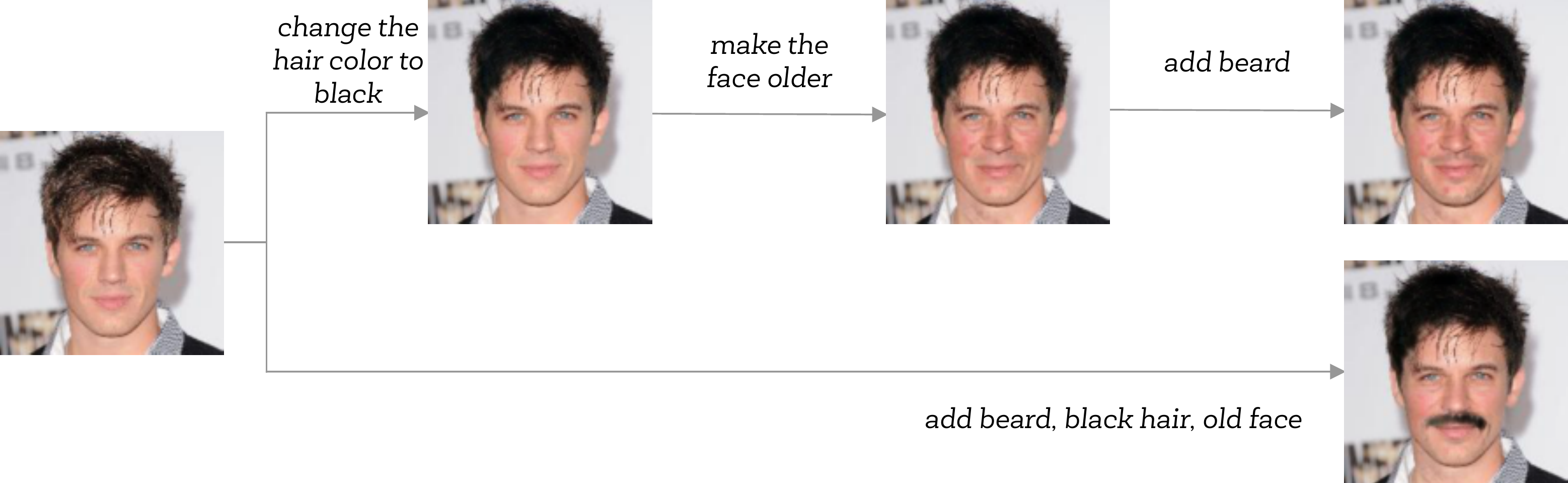} \\
    \includegraphics[width=0.85\linewidth]{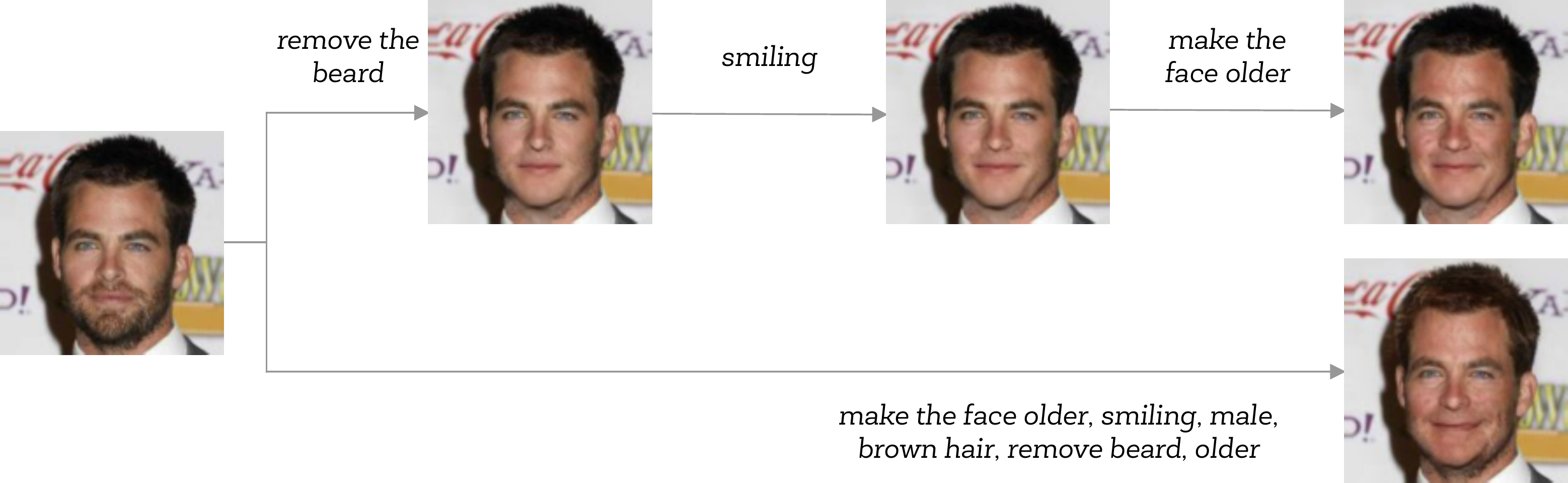} \\
	\end{tabular}
	\caption{Examples of progressive manipulation.}
	\label{fig:progressive}
\end{figure*}

\end{document}